\documentclass[lettersize,journal]{IEEEtran}
\usepackage{amsmath,amsfonts}
\usepackage{algorithmic}
\usepackage{algorithm}
\usepackage{array}
\usepackage{subcaption}
\usepackage{textcomp}
\usepackage{stfloats}
\usepackage{url}
\usepackage{verbatim}
\usepackage{cite}
\usepackage{hyperref}
\usepackage{url}
\usepackage{tabularx}
\usepackage{multirow}
\usepackage[switch]{lineno}
\usepackage{booktabs}
\usepackage{newtxtext}
\usepackage{graphicx}
\usepackage{caption}
\usepackage{longtable}
\usepackage{listings}
\usepackage{float}
\usepackage{xcolor}
\usepackage{ragged2e}
\usepackage{supertabular}
\usepackage{orcidlink}
\usepackage{xcolor}
\definecolor{red}{RGB}{153, 21, 78}
\usepackage{longtable}
\usepackage{pdflscape}
\usepackage[flushleft]{threeparttable}
\newcolumntype{L}{>{\raggedright\arraybackslash}X}

\begin{document}

\title{Generative Data Imputation for Sparse Learner Performance Data
Using Generative Adversarial Imputation Networks}

\author{Liang Zhang \orcidlink{0009-0002-0017-2569}, Jionghao Lin \orcidlink{0000-0003-3320-3907}, John Sabatini \orcidlink{0000-0002-0292-2039}, Diego Zapata-Rivera \orcidlink{0000-0002-0620-7622}, Carol Forsyth \orcidlink{0000-0003-4830-5156}, Yang Jiang \orcidlink{0000-0002-2195-5776}, John
Hollander \orcidlink{0000-0002-3270-7495}, Xiangen Hu \orcidlink{0000-0001-9045-4070}, Arthur C. Graesser \orcidlink{0000-0003-0345-6866}
\thanks{This paper is currently Under Review for IEEE Transactions on Big Data Journal. Manuscript submitted on April. 13, 2025.}
\thanks{Liang Zhang, John Sabatini, and Arthur C. Graesser are with the Institute for Intelligent Systems, University of Memphis, Memphis, TN 38152, USA (e-mail: \href{mailto:lzhang13@memphis.edu}{lzhang13@memphis.edu}, \href{mailto:jpsbtini@memphis.edu}{jpsbtini@memphis.edu}, \href{mailto:art.graesser@gmail.com}{art.graesser@gmail.com}).}
\thanks{Jionghao Lin is with the Faculty of Education, The University of Hong Kong, Hong Kong, PR China, also with the the Human-Computer Interaction Institute, Carnegie Mellon University, Pittsburgh, PA 15213, USA, and also with the Centre for Learning Analytics, Faculty of Information Technology, Monash University, Clayton, VIC 3800, Australia (e-mail: \href{mailto:jionghao@hku.hk}{jionghao@hku.hk}).}
\thanks{Diego Zapata-Rivera, Carol Forsyth and Yang Jiang are with Educational Testing Service, Princeton, NJ 08540, USA (e-mail: \href{mailto:dzapata@ets.org}{dzapata@ets.org}, \href{mailto:cforsyth@ets.org}{cforsyth@ets.org}, \href{mailto:yjiang002@ets.org}{yjiang002@ets.org}).}
\thanks{John
Hollander is with Department of Psychology and Counseling, Arkansas State University, Jonesboro, AR 72701 (e-mail: \href{mailto:jhollander@astate.edu}{jhollander@astate.edu}).}
\thanks{Xiangen Hu is with the Department of Applied Social Sciences, Hong Kong Polytechnic University, Hong Kong, PR China (e-mail: \href{mailto:xiangen.hu@polyu.edu.hk}{xiangen.hu@polyu.edu.hk}).}
}

\markboth{Journal of \LaTeX\ Class Files,~Vol.~18, No.~9, September~2020}%
{Shell \MakeLowercase{\textit{et al.}}: A Sample Article Using IEEEtran.cls for IEEE Journals}


\maketitle
\begin{abstract}
As learners engage with Intelligent Tutoring Systems (ITSs) by responding to a series of questions, their performance data, such as correct or incorrect responses, is crucial for assessing and predicting their knowledge states through analysis and modeling. However, data sparsity, often arising from skipped or incomplete responses, poses challenges for accurately assessing learning and delivering personalized instruction. To address this, we propose a generative data imputation method based on Generative Adversarial Imputation Networks (GAIN) to complete missing learning performance data. Our approach employs a three-dimensional (3D) framework structured by learners, questions, and attempts, with an adaptable design along the attempts dimension to manage varying sparsity levels. Enhanced by convolutional neural networks in the input and output layers and optimized with a least squares loss function, our GAIN-based method aligns the input and output shapes with the dimensions of question-attempt matrices across the learners' dimension. Extensive experiments on datasets from three types of ITSs, including AutoTutor Adult Reading Comprehension (ARC), ASSISTments and MATHia, demonstrate that our approach generally outperforms baseline methods, e.g., tensor factorization-based methods and other Generative Adversarial Network (GAN) variants, in imputation accuracy across different setting of maximum attempts. Bayesian Knowledge Tracing (BKT) modeling further validates the imputed data’s efficacy by estimating learning parameters, including initial knowledge \(P(L_0)\), learning rate \(P(T)\), guess rate \(P(G)\), and slip rate \(P(S)\). Results reveal that the imputed data not only enhances model fit but also closely aligns with the original sparse distributions by capturing underlying learning behaviors, indicating greater reliability in learner assessments. Kullback-Leibler (KL) divergence measurements of all these learning parameters confirm that the imputed data effectively preserve essential learning characteristics, maintaining low divergence across parameters for most datasets and attempts. These findings highlight GAIN's potential as a reliable tool for data imputation in ITSs, offering an effective solution for mitigating data sparsity issues and supporting adaptive, individualized instruction in AI-driven education. The improvements in learner modeling accuracy facilitated by GAIN-based imputation ultimately pave the way for more accurate, responsive ITS applications capable of fostering improved learning outcomes. 
\end{abstract}


\section{Introduction}
\begin{figure*}[ht!]
    \centering
\includegraphics[width=0.8\textwidth]{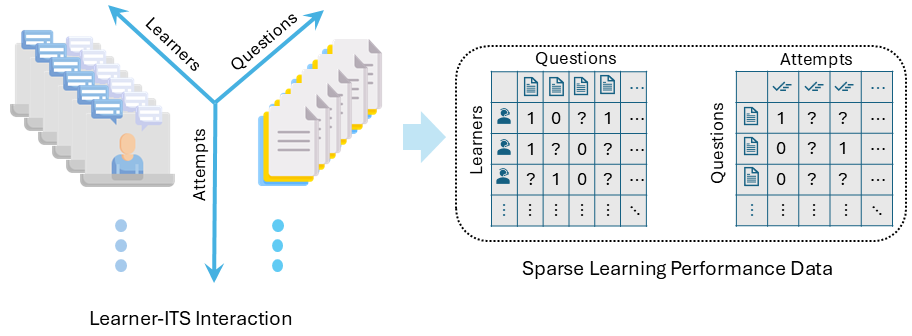}
    \caption{An illustration for sparse learning performance data in Intelligent Tutoring Systems.}
    \label{fig:sparsedata} 
\end{figure*}

\IEEEPARstart{A}{dvancements} in AI-driven technologies have significantly enhanced modern education through personalized tutoring and adaptive learning strategies on online platforms \cite{crockett2017predicting,hwang2020vision}. Intelligent Tutoring Systems (ITSs) exemplify this progress by leveraging advanced machine learning and natural language processing models to create interactive learning environments that improve outcomes across domains like literacy \cite{graesser2016reading}, mathematics \cite{heffernan2014assistments}, language learning \cite{tafazoli2019intelligent}, biology \cite{d2012gaze} and other STEM fields \cite{feng2021systematic}. As human learners interact with ITSs, often through question-and-answer scenarios with immediate responses, their performance data becomes crucial for learner modeling, enabling systems to track progress, predict future performance, and adapt instruction accordingly \cite{eglington2022optimize}. Learner models like Bayesian Knowledge Tracing (BKT) and other knowledge tracing variants utilize the learner performance data to uncover learning characteristics, estimate knowledge states and acquisition \cite{yudelson2013individualized}. However, in real-world scenarios, missing learner performance data is prevalent due to factors, such as learner dropout or disengagement \cite{chen2018disengagement}, technical issues or incomplete data logging \cite{saarela2017automatic}, biased sampling within experimental groups \cite{greer2016evaluation}, and more. These challenges often lead to sparse data, where items (i.e., questions or problems) remain unattempted (e.g., learners may bypass the question, leave it unanswered due to a lack of response initiation, or make no attempt to engage with it), alongside limited learner interactions \cite{pandey2019self, lee2022contrastive}. As shown in Figure \ref{fig:sparsedata}, missing performance records can occur along both the attempt and question dimensions during learner-ITS interactions. In the right portion of the figure’s two matrices, entries marked with ``\(?\)'' indicate missing data. This data sparsity may arise either randomly or non-randomly over the course of learner-ITS engagement. These data gaps hinder the effectiveness of learner modeling and the adaptive delivery of tailored instruction, ultimately reducing the accuracy and specificity of support provided by ITSs. For instance, sparse learning performance data can lead to biased or overfitted learner models that fail to capture a learner’s trajectory or yield misleading predictions about future performance \cite{pandey2019self,wang2019deep,lee2022contrastive,wang2023graphca}. Such misrepresentations can have adverse effects, particularly when these models inform instructional decisions or guide personalized learning paths in ITSs. To address these challenges, this study focuses on data imputation techniques for sparse learner's performance data, aiming to enhance the robustness and adaptability of ITSs in delivering personalized learning experiences within AI education.  

Various methods have been proposed to address sparse learner performance data. One traditional method, excluding incomplete observations as outliers, may seem straightforward in educational or psychological contexts \cite{saarela2017automatic}. However, this method risks discarding valuable information, exacerbating data sparsity, and undermining dataset validity. Alternatively, reverting to real-world experiments can be time-consuming, labor-intensive, and hard to replicate at scale. In light of these challenges, computational data imputation methods has gained prominence, particularly in AI research. Grounded in Rubin’s foundational principles \cite{rubin1976inference}, these computational imputation methods, such as indicator or mean imputation \cite{donders2006gentle}, regression imputation \cite{zhang2016missing}, multiple imputation \cite{rubin1978multiple}) fill missing values based on observed data patterns. While cost-effective and well-established, these methods often simplify the complexities inherent in missing data, risking bias in the resulting models \cite{batista2003analysis,donders2006gentle,seaman2012multiple,zhang2016missing}. For instance, indicator or mean imputation can introduce bias by oversimplifying the intricacies of missing data \cite{batista2003analysis,donders2006gentle}, regression imputation frequently falls short of reflecting the complete range of the underlying data structure \cite{zhang2016missing}, and multiple imputation struggles with high-dimensional correlations \cite{seaman2012multiple}. In the context of human learning data, these challenges are further compounded by its multidimensional nature. Newell and Simon's \cite{newell1972human} landmark work conceptualized human learning within a three-dimensional space, comprising the task dimension (different task environments), the performance-learning-development dimension (linking activities to timescales), and the individual-difference dimension (accounting for learner variability), capturing the dynamic and evolving nature of human cognition. Learner responses and problem-solving attempts are sequential, with past performance influencing future actions \cite{pavlik2021logistic}, variation in learner attempts \cite{zhang2023exploring}, and interdependencies among different knowledge components \cite{zhao2020modeling}. These dynamic shifts in knowledge states \cite{yudelson2013individualized}, and the multidimensional cross-effects arise from interactions between learners, tasks, and sequential attempts (e.g., a learner’s improved performance on a foundational question enhancing their ability to tackle subsequent, more complex questions, or repeated attempts on a specific task leading to knowledge reinforcement that generalizes to related tasks) \cite{zhang2024data}, introduce temporal and contextual complexities that traditional computational imputation methods struggle to address. Effectively handling sparse learning performance data in ITS contexts remains a significant challenge due to these complexities. 

Generative Adversarial Networks (GANs) have demonstrated impressive capabilities in initial image generation \cite{goodfellow2020generative}, as well as in speech and voice recognition \cite{saito2017statistical}, multimodal conversations \cite{xu2018attngan}, and beyond. By pairing a generator that produces synthetic samples with a discriminator that evaluates their authenticity, GANs leverage game-theoretic principles to learn complex data patterns and distributions, thereby enabling highly effective synthetic data generation \cite{goodfellow2014generative}. Building on these strengths, GANs have been adapted for data imputation tasks, including image inpainting for 2D images \cite{zeng2021cr}, 3D surfaces \cite{hernandez20243d}, and time series data \cite{luo2018multivariate}. One prominent model, Generative Adversarial Imputation Nets (GAIN), enhances GAN-based imputation by conditioning the generator on observed data and introducing a hint mechanism to help the discriminator detect missing patterns \cite{yoon2018gain,zhang2023systematic}, outperforming traditional methods such as MICE (Multiple Imputation by Chained Equations, a statistical method that iteratively models each variable with missing values using regression models) and missForest (a machine learning method that employs random forests to iteratively predict and impute missing data) for clinical datasets \cite{dong2021generative,yoon2018gain}. Moreover, GAIN’s context-awareness aligns with Rubin’s Rules for valid imputations, generating plausible outputs that blend seamlessly with neighboring regions \cite{zeng2021cr}. Despite its success elsewhere, GAIN’s potential to impute missing data in sparse learning performance datasets within ITSs remains unexplored. Learning performance data present unique challenges due to their multidimensional nature, encompassing individual learners, problem-solving attempts, and dynamic interactions across multiple tasks, making conventional imputation methods insufficient. A promising strategy involves adapting GAIN to three-dimensional representations of learners, items (e.g., questions), and temporal factors (e.g., time or attempts) \cite{zhang20243dg,zhang2024data}, inspired by tensor-based methods in learning engineering \cite{thai2011matrix,thai2012factorization,sahebi2016tensor}. Such a 3D approach not only captures the multifaceted nature of human learning but also allows GAIN to manage complex data structures, thereby generating more accurate imputed values \cite{zhang20243dg,zhang2024data}. Accordingly, this study investigates how GAIN can be refined for imputation in sparse learning performance datasets, optimizing its applicability to educational contexts and potentially improving overall imputation accuracy in ITSs. We are guided by the following two \textbf{R}esearch \textbf{Q}uestions:
\begin{itemize}
    \item \textbf{RQ1}: How effectively does the GAIN-based method impute sparse learning performance data in ITSs compared to established baselines?
    \begin{itemize}
        \item \textbf{RQ1.1}: How does GAIN perform across different attempts, reflecting varying levels of data sparsity?
        \item \textbf{RQ1.2}: What is the impact of GAIN on the accuracy of imputed data across various ITS datasets?
    \end{itemize}
    
    \item \textbf{RQ2}: How do the imputed data align with the original sparse learning performance in ITSs?
    \begin{itemize}
        \item \textbf{RQ2.1}: To what extent does data imputation affect learner modeling? 
        \item \textbf{RQ2.2}: How well do the imputed learning features align with those of the original sparse data? 
    \end{itemize}
\end{itemize}

The impacts of our study are twofold and extend further. First, we aim to advance the accuracy of data imputation in ITSs by applying GAIN within a 3D framework, effectively capturing the multidimensional complexities of learning behaviors that traditional methods often overlook. This advancement will significantly improve learner modeling, predictive analytics, and the ability to provide targeted, data-driven interventions in ITSs. Second, by addressing data sparsity through generative imputation, the study will enhance the adaptive capabilities of ITSs, enabling these systems to deliver more personalized and effective learning experiences for human learners. Beyond these immediate benefits, the integration of generative imputation methods is expected to establish a foundation for scalable and robust AI-driven educational tools, advancing progress tracking, assessment accuracy, and adaptive instructional designs in diverse educational contexts. Our code and results can be found at the following GitHub link: \href{ https://github.com/LiangZhang2017/GenerativeDataImputation}{https://github.com/LiangZhang2017/GenerativeDataImputation}. 

\section{Related Work}
This section reviews related work on data imputation methods in ITSs, identifying key challenges and limitations while also highlighting potential opportunities for advancement. 

\subsection{AI-driven Imputation for Sparse Learning Performance in ITSs}

Addressing data sparsity in ITSs has become a critical focus in AI-driven education, with various studies exploring imputation techniques to handle incomplete learning performance data. AI-based methods, such as deep learning frameworks, attention mechanisms, and other machine learning approaches, have played a crucial role in modeling the learning process and mitigating the effects of sparse data. For instance, Chen et al. \cite{chen2018prerequisite} utilized prerequisite concept maps to model logical relationships between knowledge concepts, enhancing knowledge state prediction in sparse data scenarios. Lu et al. \cite{lu2022cmkt} extended this by incorporating ordering pairs into concept maps. Pandey et al. \cite{pandey2019self} employed self-attention mechanisms to predict learner performance by weighting relevant prior answers. Wang et al. \cite{wang2019deep} integrated question-concept hierarchies into a deep learning framework to better model learner interactions despite data sparsity. However, challenges remain, including the labor-intensive mapping of knowledge concepts \cite{novak2006theory}, limited consideration of temporal learning dynamics \cite{thai2012factorization}, and disruption of sequential learning effects critical to knowledge organization \cite{conway2001sequential}. These gaps highlight the need for more robust approaches to addressing data sparsity in ITSs. 

\subsection{Tensor-based Imputation for Complex Learning Data}

As sparse learner-performance data become increasingly common within ITSs, the need for robust imputation methods has become essential. Tensor-based imputation has emerged as a prominent technique due to its ability to preserve the multidimensional structure of learning data and maintain intrinsic relationships across dimensions such as learners, questions, time or attempts \cite{sahebi2016tensor,zhang2023exploring}. Tensor-based approaches evolved from two-dimensional matrix factorization techniques initially applied in recommendation systems for adaptive navigation in online learning \cite{tang2005smart, liang2006courseware}. Thai-Nghe et al. \cite{thai2011factorization} extended this to three-dimensional tensor factorization, incorporating temporal effects to predict learner performance across dimensions like learners, time, and steps. They further demonstrated its potential for imputing unobserved performance in sparse datasets \cite{thai2012factorization}. Sahebi et al. \cite{sahebi2016tensor} introduced Feedback-Driven Tensor Factorization (FDTF), integrating sequences of students, quizzes, and attempts, improving knowledge representation and prediction. Later, Doan and Sahebi \cite{doan2019rank} proposed Ranked-Based Tensor Factorization (RBTF), accounting for concept forgetting and biases to promote positive learning trajectories. Zhao et al. \cite{zhao2020modeling} advanced Multi-View Knowledge Modeling (MVKM), applying tensor factorization to analyze multiple materials within a shared latent space while addressing forgetting through rank-based constraints. These and other methods \cite{sahebi2014parameterized, thai2015multi, wen2019iterative, wang2021knowledge} have enhanced predictive accuracy and addressed data sparsity in educational applications. 

These advances highlight the effectiveness of tensor-based representations in predicting and imputing missing data within sparse tensor spaces, preserving the sequential nature of learning events \cite{thai2011factorization,luo2020deep}. This approach is crucial for accurately tracing and predicting learner performance, as well as enabling efficient decomposition of interactions across multiple dimensions \cite{balavzevic2019tucker,doan2019rank}. Aligned with Rubin’s Rules on interdependence in data imputation and proven effective in recommendation systems, tensor-based techniques can uncover performance similarities and dependencies among learning events. As a result, they hold significant promise for improving imputation in educational data mining, especially when dealing with sparse datasets in ITSs. 

\subsection{Generative Data Imputation for Sparse Learning Performance}

Recent advances in generative data imputation, utilizing reconstruction mechanisms based on existing data, have achieved significant success in reducing data sparsity and outperforming traditional methods \cite{yoon2018gain,dong2021generative}. Morales-Alvarez et al. \cite{morales2022simultaneous} integrated structured latent spaces with graph neural networks, outperforming baselines such as MICE and missForest on real-world mathematics data from Eedi\footnotemark[1]\footnotetext[1]{\footnotesize Eedi Website: \url{https://eedi.com}} (an online education platform offering personalized math learning through diagnostic assessments and tailored study plans). Ma et al. \cite{ma2021identifiable} used deep generative models to impute multiple-choice question data, handling over 70\% missing rates in also Eedi’s mathematics dataset. Zhang et al. \cite{zhang20243dg} explored GAN and GPT for data augmentation in adult reading comprehension. Ongoing research continues to uncover further applications of generative imputation in learning engineering and science. 

These developments set the stage for applying more advanced models like GAIN, which preserve the multidimensional structure of learning data under a tensor format. By capturing complex relationships across dimensions, integrating GAIN within a tensor-based representation could significantly improve imputation \cite{zhang2024generative}. Inspired by these insights, our study adapts GAIN within a tensor framework to facilitate more effective imputation of sparse learning performance data in ITSs. 

\section{Methods}
This section presents our proposed data imputation method, specifically designed to address sparse learner performance data by leveraging the hierarchical relationships among learners, questions, and attempts. The framework structures the learning performance data into a 3D tensor where entries represent binary performance outcomes (correct or incorrect). Building on this 3D tensor representation, the method employs GAIN with Convolutional Neural Networks (CNNs) to effectively fill in missing values while preserving multidimensional learning dynamics and improving imputation accuracy across varying levels of data sparsity. 

\subsection{3D Tensor Representation of Sparse Learning Performance Data}

Consider an intelligent learning scenario within ITS, where a set of \(U\) learners, represented as \(\{l_1,l_2,l_3,\cdots,l_U\}\), engage with a sequence of \(N\) questions, denoted by \(\{q_1,q_2,q_3,\cdots,q_N\}\). Each question allows up to \(M\) attempts, represented by \(\{t_1,t_2,t_3,\cdots,t_M\}\), for submitting responses for answers. As learning progresses, learners' performance evolves dynamically based on the sequence of questions and repeated attempts, while the presence of missing data contributes to the sparsity of performance records. The sparse learning performance data can be represented as as 3D tensor \(\boldsymbol{\mathcal{T}}_{sparse}\in [\boldsymbol{\tau}_{uij}]^{U\times N \times M }\) or \([0,1,NaN]^{U\times N \times M }\), where each element \(\tau_{uij}\) corresponds to the observed performance of the \(u^{th}\) learner on the \(j^{th}\) question at the \(i^{th}\) attempt. Specifically, \({\tau}_{uij}\) takes the value of 1 for correct answer, 0 for incorrect answer, and \(NaN\) for unobserved data at a specific question and attempt. 

\subsection{The Proposed GAIN-based Imputation Architecture}

Building upon the basic GAN method,  we demonstrate how GAIN can be adapted to impute tensor-based learning performance data, transforming \(\boldsymbol{\mathcal{T}}_{sparse}\) into \(\boldsymbol{\mathcal{T}}_{dense}\).

\begin{figure*}
\includegraphics[width=0.83\textwidth]{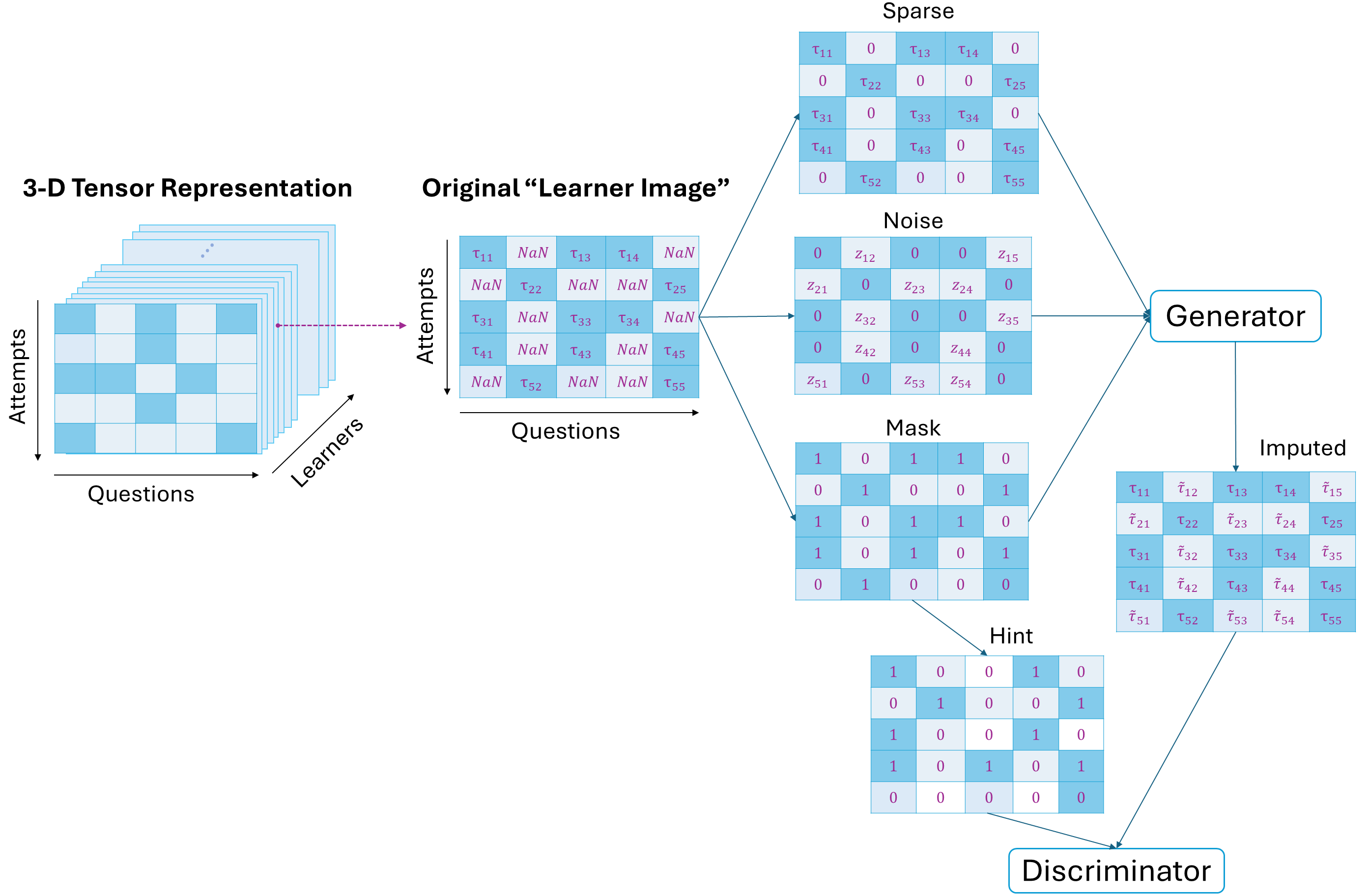}
\centering
\caption{The proposed GAIN-based imputation architecture for sparse learning performance.} \label{fig:gain}
\end{figure*}

Consider the \(\boldsymbol{\mathcal{T}}_{sparse}\), representing the learning performance of all learners. This tensor comprises layers along the learner dimension, represented as \(\boldsymbol{\mathcal{T}}_{sparse}=(\mathcal{T}_{l_{1}},\mathcal{T}_{l_{2}}, \cdots, \mathcal{T}_{l_{n}})\). Each layer, akin to a single-channel ``learner image'', is a matrix that encapsulates performance values across different questions and attempts for an individual learner. This is visualized in Figure~\ref{fig:gain}. 

For each matrix-based layer \(\mathcal{T}_{l} \in (\mathcal{T}_{l_{1}},\mathcal{T}_{l_{2}}, \cdots, \mathcal{T}_{l_{n}})\), each entry \(\tau_{lij}\) in the \(N \times M\) matrix may include the performance values of 0, 1 or \(NaN\) to present the observed data and unobserved data, respectively. One mask matrix \(\mathcal{T}_{l_{mask}}\) is supposed to map the observed and unobserved entries within the matrix \(\mathcal{T}_{l}\), with 1 signifying observed data, and 0 indicates unobserved data. One noise matrix \(\mathcal{Z}\) with dimensions matching \(\mathcal{T}_{l}\), is initialized. These matrices collectively function as inputs to the generator in the GAIN architecture, producing the output \(\mathcal{T}_{lG}=G(\mathcal{T}_{l},\mathcal{T}_{l_{mask}}, (1-\mathcal{T}_{l_{mask}}) \odot \mathcal{Z})\) \cite{yoon2018gain}. Here, the \(\odot\) denotes as Hadamard product, indicating element-wise multiplication. The imputed matrix \(\mathcal{T}_{l_{imputed}}= \mathcal{T}_{l_{mask}} \odot \mathcal{T}_{l} + (1-\mathcal{T}_{l_{mask}}) \odot \mathcal{T}_{lG}\), effectively merges observed and generated data to fill in unobserved entries. Particularly, a hint matrix \(\mathcal{T}_{l_{hint}}\), also matching the dimensions of \(\mathcal{T}_{l}\) and derived from the mask matrix \(\mathcal{T}_{l_{mask}}\), is introduced. It employs a hint rate to specify the conditional probability that a specific entry in \(\mathcal{T}_{l_{imputed}}\) can be observed, given both \(\mathcal{T}_{l_{imputed}}\) and \(\mathcal{T}_{l_{hint}}\). Thereby, the discriminator within the GAIN architecture, formulated as \(D(\mathcal{T}_{l_{imputed}}, \mathcal{T}_{l_{hint}})\), evaluate this as probability \cite{yoon2018gain}. We train \(D(\cdot)\) to maximize the probability of correctly predicting the \(\mathcal{T}_{l_{mask}}\), while the \(G(\cdot)\) is trained to minimize the likelihood of \(D(\cdot)\) correctly predicting \(\mathcal{T}_{l_{mask}}\). So, we introduce the objective function \(V(D,G)\) \cite{yoon2018gain}: 
\begin{equation}
\footnotesize
\begin{gathered} \label{eq:vali}
    V(D,G) = E\Big[ \mathcal{T}_{l_{mask}}^{T} \log D(\mathcal{T}_{l_{imputed}}, \mathcal{T}_{l_{hint}}) + \\
    (1 - \mathcal{T}_{l_{mask}})^{T} \log(1 - D(\mathcal{T}_{l_{imputed}}, \mathcal{T}_{l_{hint}})) \Big]
\end{gathered}
\end{equation}

Our proposed imputation architecture incorporates several novel modifications and configurations from the initial GAIN architecture \cite{yoon2018gain}. See below for further details: 
\begin{itemize}
    \item The convolutional layers are employed for both the generator and discriminator, diverging from the original architecture' reliance on dense layers. Five convolutional neural network (CNN) layers \cite{lecun1989handwritten}, excluding the input and output layers, with the ReLU activation function are applied to the output of each layer. 
    \item During the iterative training phase, the observed data from \(\mathcal{T}_{l}\) and the corresponding imputed data from \(\mathcal{T}_{lG}\) are utilized for optimization via the least square loss function, specifically the Root Mean Square Error (RMSE). This method is chosen to not only ensure enhanced stability and superior quality of the generated data \cite{mao2017least,yoon2020gamin} but also align with probability-based predictions of learning performance in peer research on ITSs \cite{yudelson2013individualized,gervet2020deep,pavlik2021logistic}. 
    \item By incorporating a reshape function in the generator's output layer, the shape of generated data \(\mathcal{T}_{lG}\) is flexible adjustment to fit the given ``learner image'' shape, thus accommodating variations across different lesson scenarios without being constrained to a fixed shape, as commonly seen in image-oriented research \cite{goodfellow2014generative,chen2016infogan,wang2021pc}. 
\end{itemize} 

The theoretical foundation underlying the inference logic and model assumptions in our proposed generative data imputation approach encompasses the following aspects: 
\begin{itemize}
    \item \textbf{Inference Logic.} The entry set within \(\boldsymbol{\mathcal{T}}_{sparse}\) can be categorized into two subsets: \(\mathcal{T}_{observed}\) for existing values (0 and 1) and \(\mathcal{T}_{unobserved}\) for missing ones (\(NaN\)). The inference model, formulated as \(f_{impute}(\mathcal{T}_{unobserved}|\mathcal{T}_{observed})\), is principle for data imputation,  leveraging observed data patterns to impute missing values and predict outcomes \cite{rubin1976inference}.
    \item \textbf{Model Assumptions.} Our imputation model operates under several key assumptions within a tensor-based framework: \textbf{(a)} \textit{Probability-based prediction}: Assumes predicted learning performance is a continuous probability between 0 and 1, indicative of knowledge mastery \cite{baker2008more}. \textbf{(b)} \textit{Latent domain knowledge relations}: Posits that unobservable latent relationships within the domain knowledge implicitly influence knowledge mastery \cite{corbett1994knowledge,essa2016possible}. \textbf{(c)} \textit{Similarity in learning for individual learners}: Suggests a shared relevance and usefulness of knowledge among learners, aiding in predicting knowledge mastery \cite{thai2011factorization,thai2011matrix}. \textbf{(d)} \textit{Performance interactions influenced by sequence effects}: Acknowledges that learners' interactions with sequential questions are shaped by priming and recency effects, affecting comprehension and performance \cite{conway2001sequential,ramscar2016learning}. \textbf{(e)} \textit{Maximum attempt assumption}: Defines an empirical maximum number of attempts a learner might require (based on the experimental dataset), highlighting the importance of assessing comprehensive learning states through repeated trials \cite{corbett1994knowledge}. 
\end{itemize} 

\section{Experiments}
In this section, we first introduce the experimental setup, including the evaluation datasets, baselines, and settings. Then,
we evaluate the proposed GAIN model and compare it with the baseline models. Finally, we conduct experiments to illustrate the impact of the important learning parameters (derived from BKT) in learning performance.

\subsection{Dataset}
\begin{table*}[ht!] 
\centering
\footnotesize
\renewcommand{\arraystretch}{1}  %
\caption{Dataset for the ARC AutoTutor, ASSISTments and MATHia lessons.} 
\label{tab:all_dataset}
\begin{tabularx}{\textwidth}{>{\raggedright\arraybackslash\hsize=0.7\hsize}X >{\raggedright\arraybackslash\hsize=1.4\hsize}X >{\centering\arraybackslash\hsize=.5\hsize}X >{\centering\arraybackslash\hsize=.5\hsize}X >{\centering\arraybackslash\hsize=.5\hsize}X}
\toprule
Dataset & Lesson Topics & \#Learners & \#Questions & \#Attempts \\ \midrule 
ARC Lesson 1 & Cause and Effect & 118 & 9 & 9 \\
ARC Lesson 2 & Problems and Solution & 140 & 11 & 5 \\
ASSISTments Lesson 1 & Algebra Symbolization Studies & 318 & 64 & 4 \\
ASSISTments Lesson 2 & Skill Builder & 392 & 20 & 4
\\
MATHia Lesson 1 & Scale Drawings & 500 & 28 & 4 \\
MATHia Lesson 2 & Analyzing Models of Two-Step Linear Relationships & 500 & 6 & 4 \\
\bottomrule 
\end{tabularx}
\end{table*}

To fully evaluate the performance of our revised GAIN model adapted for tensor-based learning data imputation, we utilized three datasets from distinct lessons across different ITSs: AutoTutor ARC lessons\footnotemark[1]\footnotetext[1]{\footnotesize AutoTutor Moodel Website: \url{https://sites.autotutor.org/}; Adult Literacy and Adult Education Website: \url{https://adulted.autotutor.org/}}, ASSISTments\footnotemark[2] \footnotetext[2]{ASSISTments Website: \url{https://new.assistments.org/}} and the MATHia\footnotemark[3]\footnotetext[3]{MATHia Website: \url{https://www.carnegielearning.com/solutions/math/mathia/}} dataset from mathematics class. As shown in Table \ref{tab:all_dataset}, the AutoTutor ARC lessons include topics such as ``\textit{Cause and Effect}'' (ARC Lesson 1) and ``\textit{Problems and Solution}'' (ARC Lesson 2), each consisting of 9 to 11 multiple-choice questions designed to test adults' reading comprehension. This study received ethical approval from the Institutional Review Board (IRB), approval number H15257. The ASSISTments dataset includes lessons on ``\textit{Algebra Symbolization Studies}'' (ASSISTments Lesson 1)\footnotemark[4]\footnotetext[4]{Assistments 2008-2009: \url{https://pslcdatashop.web.cmu.edu/DatasetInfo?datasetId=388}} and ``\textit{Skill Builder}'' (ASSISTments Lesson 2)\footnotemark[5]\footnotetext[5]{Assistments 2012-2013:\url{https://sites.google.com/site/assistmentsdata/datasets/2012-13-school-data-with-affect?authuser=0}}. The MATHia dataset\footnotemark[6]\footnotetext[6]{MATHia 2019-2020: \url{https://pslcdatashop.web.cmu.edu/Project?id=720}} covers algebra lessons, specifically ``\textit{Scale Drawings}'' (MATHia Lesson 1) and ``\textit{Analyzing Models of Two-Step Linear Relationships}'' (MATHia Lesson 2). Table~\ref{tab:all_dataset} provides further details, including the total number of learners, questions and attempts. 

\subsection{Baselines}

Our study compares the GAIN-based imputation method with a range of baseline techniques, including methods from the tensor factorization and GAN series. A detailed description of these baseline approaches is provided below.

\textbf{Tensor Factorization:} The basic tensor factorization factorizes the sparse tensor \(\boldsymbol{\mathcal{T}}_{sparse}\) into two components: a learner latent matrix capturing abilities and learning-related features, and a latent tensor representing knowledge during question attempts \cite{zhang2023exploring,zhang20243dg}. A rank-based constraint is used to maintain a generally positive learning trend and accommodate forgetting or slipping \cite{doan2019rank}. This refined method enhances data imputation within tensor-based structures, providing a robust solution for handling sparse data. 

\textbf{CANDECOMP/PARAFAC Decomposition (CPD):} Drawing on the principle of classic CPD \cite{carroll1970analysis,harshman1970foundations}, the sparse tensor \(\boldsymbol{\mathcal{T}}_{sparse}\) is decomposed into three factor tensors that capture learner, attempt and question-related factors in a multidimensional tensor form. A rank-based constraint is additionally applied to enhance the decomposition's accuracy. 

\textbf{Bayesian Probabilistic Tensor Factorization (BPTF):} The BPTF \cite{xiong2010temporal} is employed to approximate the sparse tensor \(\boldsymbol{\mathcal{T}}_{imputed}\) through the decomposition into a sum of outer products of three lower-dimensional factor tensors. This approach leverages Bayesian inference for sampling both the factor tensors and the precision of observed entries, effectively enhancing the model’s capacity to manage data sparsity and uncertainty \cite{xiong2010temporal,morise2019bayesian}. 

\textbf{Generative Adversarial Network (GAN):} At one core of the GAN, the ``learner image'' extracted from \(\boldsymbol{\mathcal{T}}_{sparse}\) (depicted in Figure~\ref{fig:gain}), constitutes the base input for the GAN. The GAN architecture includes a generator that simulates data resembling observed entries and a discriminator that assesses the authenticity of this generated data \cite{goodfellow2014generative}. It uses a consistent CNN layer configuration and least squares loss for optimization. 

\textbf{Information Maximizing Generative Adversarial Nets (InfoGAN):} The InfoGAN \cite{chen2016infogan} enhances the traditional GAN framework by integrating the noise with two structured latent variables, allowing for the capture of salient, structured semantic features, such as those relating to learner attributes in ITSs (e.g., initial learning ability and learning rate). The generator generates imputed \(\boldsymbol{\mathcal{T}}_{imputed}\) and decodes latent variables. An auxiliary distribution improves the estimation of these variables' posterior, boosting mutual information between latent codes and observations and ensuring that the generated outcomes are meaningfully informed. 

\textbf{AmbientGAN:} AmbientGAN \cite{bora2018ambientgan} is used to impute sparse learning performance data by training on partially observed or corrupted data within a GAN framework. It incorporates a dynamically adjusted Gaussian blur in the measurement process, enabling the discriminator to effectively distinguish between real and generated data measurements and accurately infer the original dataset's true distribution. 

\subsection{Experimental Settings and Evaluations}

In our experiments for imputing sparse learning performance data, we incorporate several tailored configurations to optimize model training and evaluation. \textbf{(a)} \textit{Cross-Validation}: We employ a five-fold cross-validation strategy, repeated over five cycles, for each model to ensure consistency and reliability of the results. \textbf{(b)} \textit{Varying Attempt Settings}: To assess the stability of the models' data imputation performance across different levels of data sparsity, we evaluate them under various maximum attempt settings. \textbf{(c)} \textit{Maximum Iterations}: All models are trained for a maximum of 100 iterations to allow adequate learning while monitoring convergence. Training is stopped early if convergence is achieved before reaching the iteration limit, ensuring computational efficiency without compromising the model's performance. \textbf{(d)} \textit{Learning Rate}: We use a learning rate of either 0.0001 or 0.00001, depending on the model, to promote steady progress and convergence during training. \textbf{(e)} \textit{Regularization Techniques}: To prevent overfitting, dropout and batch normalization are integrated into the training process of the GAN-based methods. \textbf{(f)} \textit{Imputation Accuracy Evaluation Metric}: We use the Root Mean Square Error (RMSE) to evaluate the models' performance in data imputation, following previous research \cite{xiong2010temporal,sahebi2016tensor,yoon2018gain}. \textbf{(g)} \textit{Mean RMSE and Variability:} The mean RMSE values were obtained over five runs, with standard errors computed to measure variability. \textbf{(h)} \textit{Measuring Sparsity Level}: The sparsity level of the tensor-based learning performance data is computed as the percentage of missing values relative to the total number of elements in the dataset.

\subsection{Evaluating Impacts of Data Imputation on Learning Parameters Using Bayesian Knowledge Tracing}

To assess the effectiveness of our proposed data imputation method, we analyze the changes in estimated learning parameters before and after imputation using Bayesian Knowledge Tracing (BKT) \cite{corbett1994knowledge}. BKT was chosen due to its explainable parameters, which are directly tied to specific learning features in student performance data. Specifically, we focus on four core probability-based parameters in BKT \cite{pardos2010modeling}: \(P(L_0)\), the probability of initial knowledge, representing the likelihood that the learner has already mastered a skill at the start; \(P(T)\), the probability of learning rate, indicating the chance that the learner transitions from an unmastered to a mastered state for a particular skill; \(P(G)\), the probability of guess rate, reflecting the likelihood that the learner answers correctly by guessing while still in an unmastered state; \(P(S)\), the probability of slip rate, denoting the probability of an incorrect answer despite the learner being in a mastered state. We compare these parameters, derived from the imputed datasets, with those from the original sparse datasets to assess the impact of imputation on the accuracy and reliability of student modeling. The RMSE is also used to evaluate the performance of the BKT modeling, providing a quantitative measure of the impact of our proposed GAIN-based data imputation method. Additionally, significance testing using the one-sided paired test is performed to assess the statistical differences between the original and imputed datasets. 

To quantify the changes in BKT parameters before and after imputation, we calculated the Kullback-Leibler (KL) divergence, which measures the difference between two probability distributions, specifically the distributions of BKT parameters derived from the original and imputed datasets. We applied Kernel Density Estimation (KDE) to obtain smoothed probability density functions for each parameter, addressing issues of data sparsity and variance. For each BKT parameter, KDE was used to estimate the continuous probability distributions from both the original and imputed data, and the KL divergence between these two distributions was then computed using the following formula \cite{kullback1951information}:
\begin{equation}
\footnotesize
\text{KL}(P \parallel Q) = \int_{-\infty}^{\infty} P(x) \log\left( \frac{P(x)}{Q(x)} \right) \, dx
\end{equation}
where \(P(x)\) is the probability density function for learning parameters from the original data and \(Q(x)\) is that from the imputed data. This calculation provides a measure of how much the imputation alters the underlying parameter distributions, thereby evaluating the effectiveness and impact of the GAIN method using the BKT modeling. 

\section{Results}

\subsection{Data Imputation Accuracy}

\begin{figure*}[h!t]
\centering
\includegraphics[width=7.2in]{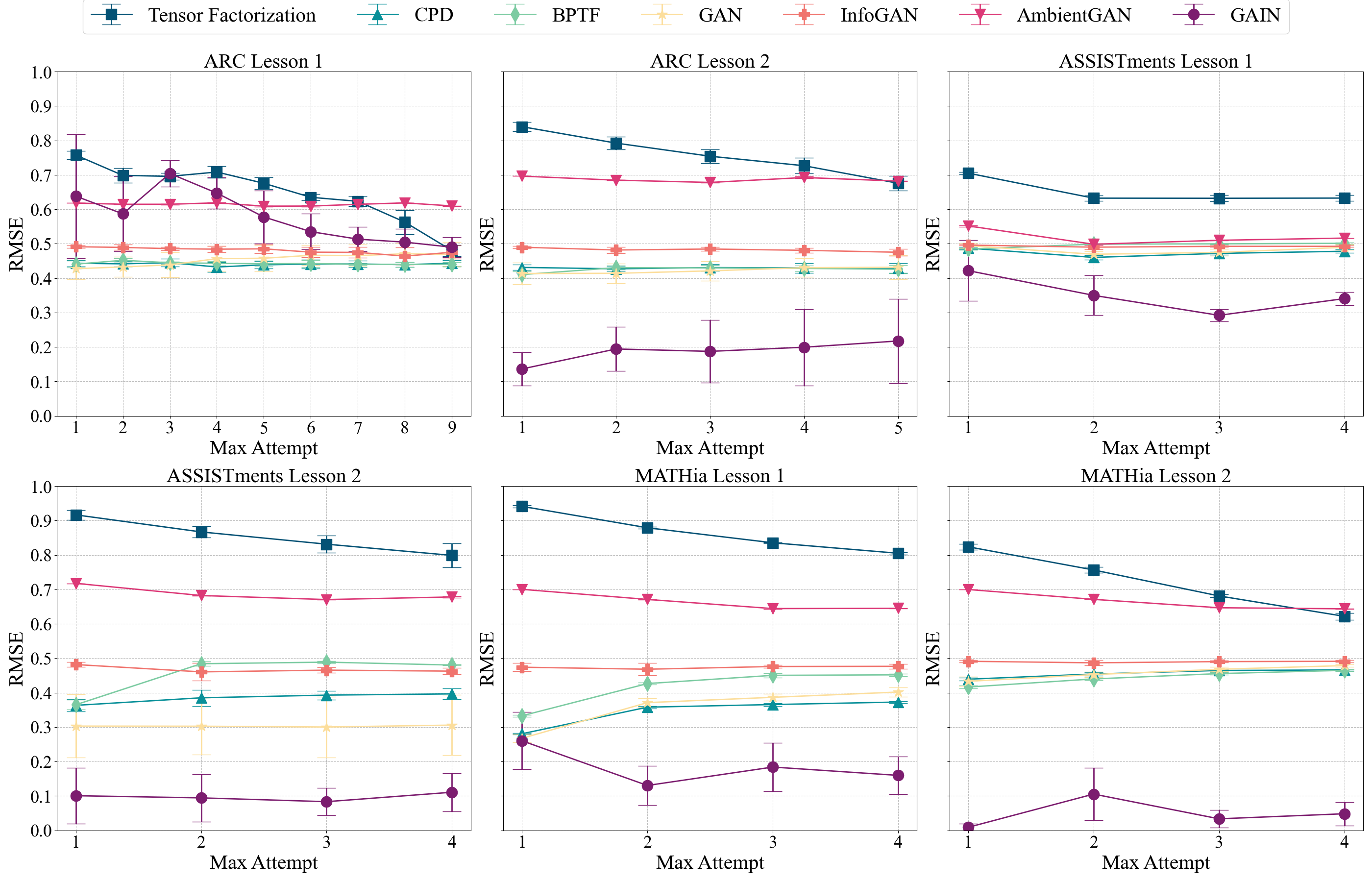}
\caption{RMSE comparisons of data imputation models across different attempts for different lessons dataset.}
\label{fig:GAIN_imputation}
\end{figure*}

\textbf{RQ1} investigates how effectively the GAIN-based method imputes sparse learning performance data in ITSs compared to established baselines. This question is examined through the following results. The imputation accuracy of various models (Tensor Factorization, CPD, BPTF, GAN, InfoGAN, AmbientGAN, and GAIN) on sparse learning performance data from six lessons across AutoTutor ARC, ASSISTments, and MATHia is evaluated using RMSE, as shown in Figure~\ref{fig:GAIN_imputation}. The figure also presents RMSE values across different Max Attempt settings (addressing \textbf{RQ1.1}), where smaller RMSE values indicate higher imputation accuracy. Error bars in the figure represent standard errors, as previously discussed. As observed, GAIN frequently achieves the lowest RMSE across most lessons, demonstrating superior imputation accuracy, particularly in the ASSISTments and MATHia datasets, where it distinctly outperforms other models (addressing \textbf{RQ1.2}). However, an exception occurs in ARC Lesson 1, where GAIN's RMSE is larger than that of BPTF, GAN, and InfoGAN, with longer error bars suggesting greater variance and reduced accuracy in imputation. This divergence likely reflects the unique characteristics of reading comprehension assessment, where questions build upon multiple interrelated skills (decoding, vocabulary knowledge, and inferential reasoning). Unlike mathematical problems where knowledge components are often more discretely defined, reading comprehension questions typically draw upon overlapping cognitive processes. This interconnected nature of reading comprehension skills may challenge GAIN's ability to accurately impute missing values, particularly when learners exhibit uneven skill profiles across different comprehension strategies. The relative higher RMSE values across different Max Attempt highlight challenges in accurately imputing this dataset. This unique case underscores complex data or model interactions that require further investigation. Additionally, while GAN performs well in ARC Lesson 1, it is slightly less robust than GAIN, with CPD and BPTF also showing competitive results. 

Despite its overall superior performance, GAIN exhibits greater variance in its results, as reflected by the longer error bars in Figure~\ref{fig:GAIN_imputation}, indicating reduced stability in its data imputation. This heightened variance suggests that while GAIN often achieves superior accuracy, its consistency may be compromised under certain data conditions, potentially requiring additional tuning or pre-processing to enhance stability. In contrast, other baseline models, such as Tensor Factorization and CPD, show lower variance, suggesting they may offer more reliable imputations in specific contexts, even though they do not always achieve the lowest RMSE for optimal imputation accuracy. 

\subsection{Comparative Analysis of BKT Modeling Accuracy Using RMSE: Original Sparse Data vs. Imputed Data}

\begin{table*}[ht!]
    \centering
    \footnotesize
    \caption{RMSE Comparisons of BKT modeling between the original and imputed data.}
    \label{tab:bkt_comparison}
    \renewcommand{\arraystretch}{1.1}
    \begin{tabularx}{\textwidth}{@{\extracolsep{\fill}}l l *{9}{c}@{}}
        \toprule
        \multirow{2}{*}{\textbf{Dataset}} & \multirow{2}{*}{\textbf{Type}} & \multicolumn{9}{c}{\textbf{Max Attempt}} \\
        \cline{3-11}
        & & 1 & 2 & 3 & 4 & 5 & 6 & 7 & 8 & 9 \\
        \midrule
        \multirow{3}{*}{\centering ARC Lesson 1} 
        & Original  & 0.395 & 0.400  & 0.395  & 0.396 & 0.396 & 0.400 & 0.397 & 0.399 & 0.398 \\
        & Imputed & 0.404 & 0.421 & 0.401 & 0.377 & 0.360 & 0.341 & 0.329 & 0.316 & 0.305  \\
        & Difference  & +0.009 & +0.021 & +0.006 & -0.019 & -0.036 & -0.059 & -0.068 & -0.083 & -0.093 \\
        \midrule
        \multirow{3}{*}{\centering ARC Lesson 2} 
        & Original  & 0.391  & 0.390  & 0.388  & 0.389 & 0.390 & --- & --- & --- & --- \\
        & Imputed & 0.423 & 0.444 & 0.403 & 0.370 & 0.344 & --- & --- & --- & --- \\
        & Difference  & +0.032 & +0.054 & +0.015 & -0.019 & -0.046 & --- & --- & --- & --- \\
        \midrule
        \multirow{3}{*}{\centering ASSISTments Lesson 1} 
        & Original  & 0.472 & 0.469 & 0.471 & 0.475 & ---  & --- & --- & --- & ---  \\
        & Imputed & 0.364 & 0.318 & 0.294 & 0.280 & --- & --- & --- & --- & --- \\
        & Difference  & -0.108 & -0.151 & -0.177 & -0.195 & --- & --- & --- & --- & --- \\
        \midrule
        \multirow{3}{*}{\centering ASSISTments Lesson 2} 
        & Original  & 0.320 & 0.398 & 0.407 & 0.417 & --- & --- & --- & --- & --- \\
        & Imputed & 0.277 & 0.204 & 0.172 & 0.154 & --- & --- & --- & --- & --- \\
        & Difference & -0.043 & -0.194 & -0.235 & -0.263 & --- & --- & --- & --- & --- \\
        \midrule
        \multirow{3}{*}{\centering MATHia Lesson 1} 
        & Original & 0.260 & 0.366 & 0.410 & 0.427 & --- & --- & --- & --- & --- \\
        & Imputed & 0.228 & 0.471 & 0.476 & 0.454 & --- & --- & --- & --- & --- \\
        & Difference  & -0.031 & +0.105 & +0.066 & +0.027 & --- & --- & --- & --- & --- \\
        \midrule
        \multirow{3}{*}{\centering MATHia Lesson 2} 
        & Original  & 0.408 & 0.428 & 0.452 & 0.466 & ---  & --- & --- & --- & --- \\
        & Imputed & 0.372 & 0.388 & 0.424 & 0.450& --- & --- & --- & --- & --- \\
        & Difference & -0.036 & -0.040 & -0.028 & -0.016 & --- & --- & --- & --- & --- \\
        \bottomrule
    \end{tabularx}
    \begin{threeparttable}
    \begin{tablenotes}
         \item[*] Note: The positive symbol ``+'' denotes an increase in the RMSE value, while the negative symbol ``-'' indicates a decrease. The symbol ``---'' signifies that the maximum number of attempts was not applicable for that particular dataset.
    \end{tablenotes}
    \end{threeparttable}
\end{table*} 

\begin{figure*}[h!t]
\centering
\includegraphics[width=7.2in]{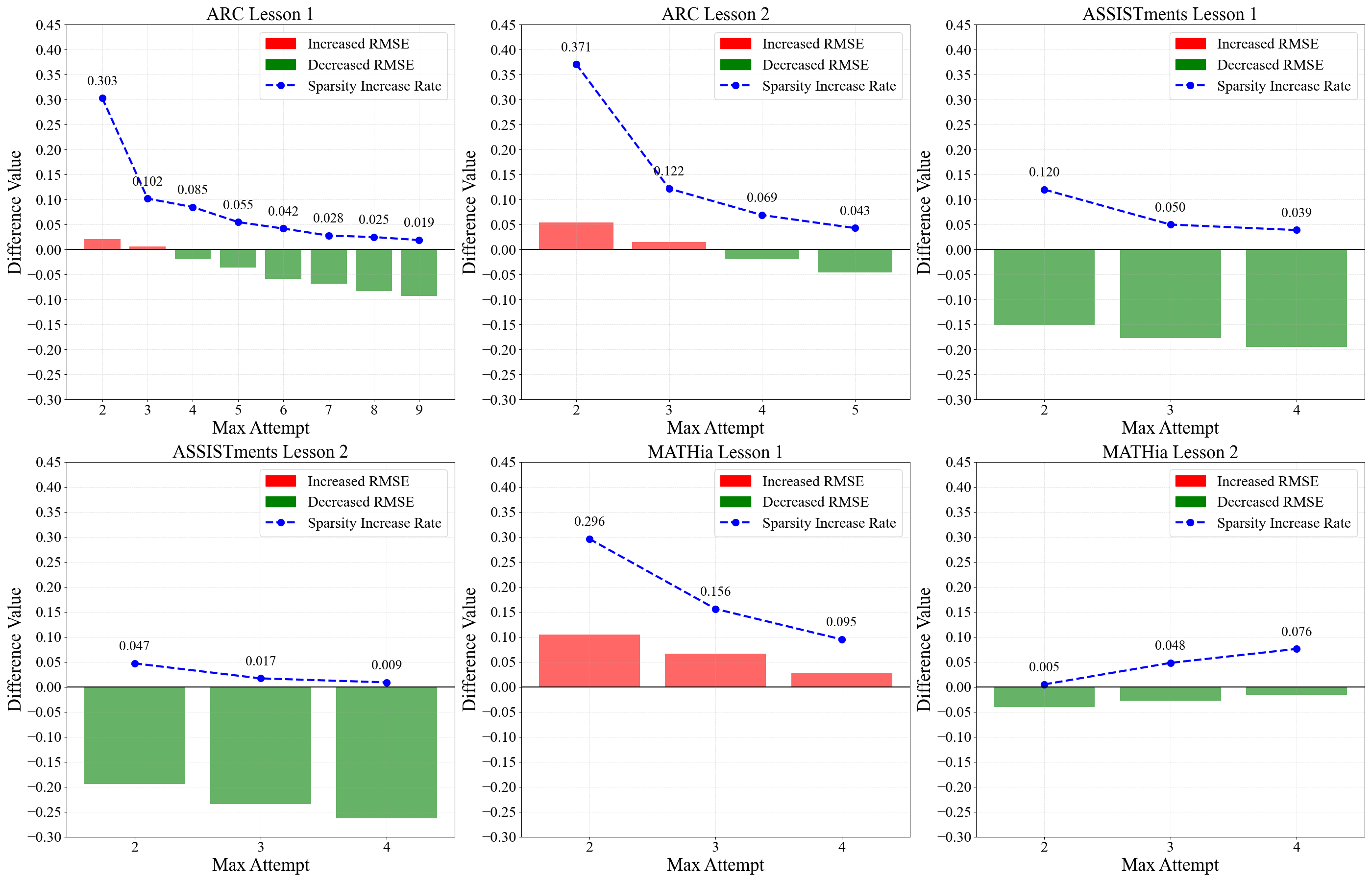}
\caption{RMSE changes with varying sparsity increase rates (corresponding to Max Attempts).}
\label{fig:rmse_sparsity}
\end{figure*} 

\textbf{RQ2} examines how the imputed data align with the original sparse learning performance in ITSs, focusing on the impact of GAIN-based data imputation on learner modeling, as exemplified by BKT (\textbf{RQ2.1}). This aspect is addressed through the following results. Table~\ref{tab:bkt_comparison} compares the BKT model's performance on original and imputed data across six lessons, highlighting RMSE differences for varying maximum attempt. For each lesson, the table presents RMSE values for the original data ("Original"), the imputed data ("Imputed"), and their difference ("Difference"), calculated as the imputed RMSE minus the original RMSE. A negative difference indicates a decrease in RMSE, reflecting improved BKT modeling accuracy after imputation. Conversely, a positive difference signifies an increase in RMSE, indicating reduced BKT modeling accuracy. For ARC Lesson 1, RMSE initially increases for Max Attempts 1 to 3, indicating a marginal decline in model performance, but decreases from Max Attempt 4 onward, suggesting improved model accuracy at higher attempts. Similarly, in ARC Lesson 2, RMSE increases at lower attempts but decreases at higher attempts. In the ASSISTments datasets, imputed data consistently show reduced RMSE across all attempts, demonstrating enhanced BKT model performance post-imputation. In contrast, for the MATHia datasets, RMSE increases in Lesson 1 for most attempts, while in Lesson 2, the imputed data consistently result in lower RMSE values. To evaluate the significance of these changes, a paired one-sided t-test was conducted comparing original and imputed RMSE values for each attempt. The t-test produced (\(t=-3.2552, p=0.0014\)), confirming that imputation significantly improves BKT model performance (addressing \textbf{RQ2.1}). The \(t=-3.2552\) indicates a substantial difference between the mean RMSE values of the original and imputed datasets, with the negative sign suggesting that the imputed data generally result in lower RMSE values. Moreover, the \(p=0.0014\), well below the threshold of 0.05, provides strong statistical evidence supporting the significance of these improvements. This statistical evidence highlights the effectiveness of GAIN-based imputation in mitigating the effects of data sparsity and reliably improving model performance across lessons and datasets. 

Additionally, a notable observation is the difference in RMSE values from BKT modeling between the original and imputed datasets. These differences, which include both increases and decreases (as shown in Table \ref{tab:bkt_comparison}), may be attributed to changes in data sparsity driven by the increasing number of attempts in the original dataset. This observation is further verified by the following analysis results, which provide additional insights into \textbf{RQ2.1}. See the Figure~\ref{fig:rmse_sparsity}, where the terms ``Increased'' and ``Decreased'' in RMSE correspond to positive and negative ``Difference'' values in Table \ref{tab:bkt_comparison}, respectively. The sparsity increase rate is calculated by dividing the sparsity level values by their increase with Max Attempt (refer to \hyperref[appendix:A]{Appendix A} for more details on how sparsity levels change with increasing Max Attempt, where the slopes of the lines with respect to Max Attempt represent the ``sparsity increase rate''). As observed from Figure~\ref{fig:rmse_sparsity}, the sparsity increase rate generally decreases across Max Attempt in most lessons, with the exception of MATHia Lesson 2, where it shows a gradual increase. In most situations with increased RMSE, the sparsity increase rate tends to be higher, whereas decreased RMSE generally corresponds to lower sparsity increase rates. For example, in ARC Lesson 1, RMSE increases at Max Attempts 2 and 3, where the sparsity increase rates are 0.303 and 0.102, respectively. In contrast, from Max Attempt 4 onward, where RMSE decreases, the sparsity increase rate drops to 0.085 or lower. Similarly, in ARC Lesson 2, RMSE increases at Max Attempts 2 and 3, with sparsity increase rates of 0.371 and 0.122, which are notably higher than those observed at subsequent attempts, where RMSE decreases and the sparsity increase rates drop to 0.069 and 0.043. In MATHia Lesson 1, RMSE increases across Max Attempts 2, 3, and 4, where all sparsity increase rates remain above 0.095. These observations suggest that steeper changes in sparsity levels, which reflect a rapid increase in the proportion of missing data with each additional attempt, make it more challenging for the imputation model to accurately predict missing values, as evidenced by the increased RMSE values. Conversely, when the sparsity increase rate is lower, the model faces fewer abrupt changes in data sparsity, allowing for more accurate imputation and resulting in decreased RMSE values. Additionally, all observed sparsity increase rates are consistently below 0.095 in cases where RMSE decreases, indicating improved GAIN-based imputation. However, it remains unclear whether this value represents a definitive threshold or merely coincides with an underlying threshold that influences the effectiveness of data imputation. Identifying such a threshold, if it exists, presents an intriguing avenue for future research to better understand the relationship between sparsity dynamics and imputation accuracy. 

\subsection{Divergence Measurement of Learning Features in Imputed Data from Original Data}

\begin{figure*}[ht!]
\centering
\includegraphics[width=7in]{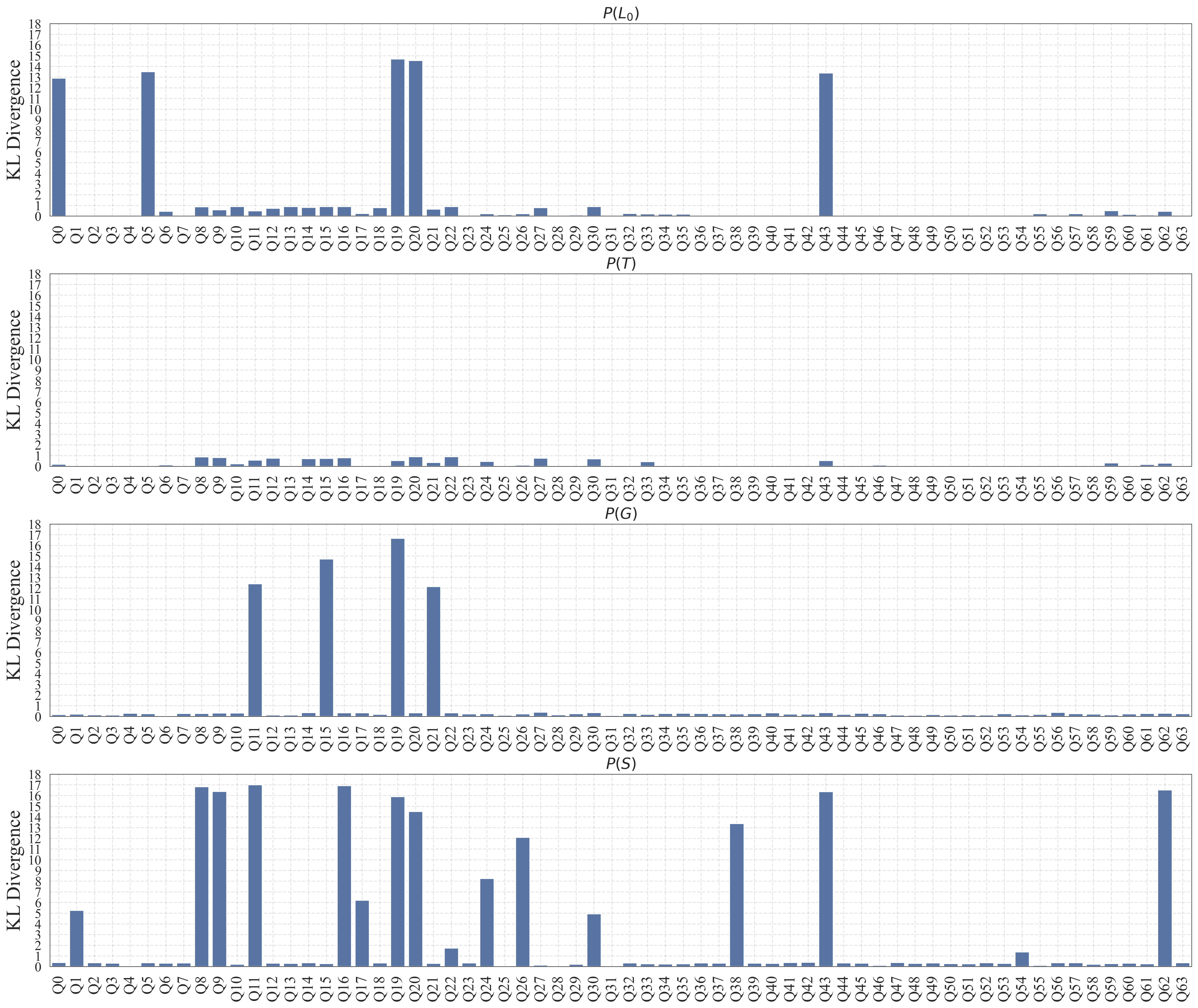}
\caption{Comparison of BKT parameters (Original vs. Imputed Learning Performance Data) for ASSISTments Lesson 1 (Max Attempt = 4).}
\label{fig:kj_comparison_example}
\end{figure*}

The KL divergence measurements for the estimated learning parameters (\(P(L_0)\), \(P(T)\), \(P(G)\), and \(P(S)\)) distributions for individual questions, derived from BKT modeling, demonstrate the effectiveness of the proposed imputation model in preserving critical learning features and characteristics in the imputed data compared to the original data. These results collectively address \textbf{RQ2.2}. As shown in Figure \ref{fig:kj_comparison_example}, the KL divergence value, calculated using relative entropy,  indicate the degree of alignment between question-wise learning parameters in the imputed and original datasets, with ASSISTments Lesson 1 (Max Attempt = 4) serving as an example. The key observations are as follows:
\begin{itemize}
    \item For \(P(L_0)\), the KL divergence values are quite low for most questions (below 1), with a few exceptions (including Q0, Q5, Q19, Q20, and Q43), which exhibit higher divergence (greater than 1). This indicates that while the imputed data generally align well with the original data in modeling initial knowledge, some questions proved more challenging to accurately impute, potentially introducing slight bias in these cases.  
    \item For \(P(T)\), the KL divergence values are consistently low, remaining below 1 for all questions. This indicates that the imputation process had minimal impact on the learning rate parameter, suggesting a strong alignment between the imputed and original data in representing learning rates. The low divergence in \(P(T)\) demonstrates the proposed GAIN model's capability in data imputation, effectively preserving the integrity of learning progression estimates across questions. 
    \item For \(P(G)\), the KL divergence values are generally low (below 1) across most questions, indicating that the imputation model closely aligns with the original data in modeling guessing behavior. This suggests that our proposed GAIN model the effectively captures overall trends in guessing. However, there are notable spikes in divergence values for a few specific questions (e.g., Q11, Q15, Q19 and Q21), where the imputed data differs significantly from the original. These cases suggest that the imputation model may introduce some bias by altering the variability in guessing behavior observed in the sparse data, potentially underestimating guessing tendencies for certain questions. 
    \item For \(P(S)\), most questions demonstrate low divergence (below 1), while approximately 15 questions show significant variability with KL divergence values exceeding 1. This indicates that, in most cases, the imputation model closely follows the original data for slip rates. However, for specific questions (e.g., Q1, Q8, Q9, Q11, Q16, etc.), higher divergence suggests that the model may introduce variability in slip characteristics, particularly in cases where the original data was sparse or noisy. 
\end{itemize}

\begin{figure*}[h!t]
\centering
\includegraphics[width=7.2in]{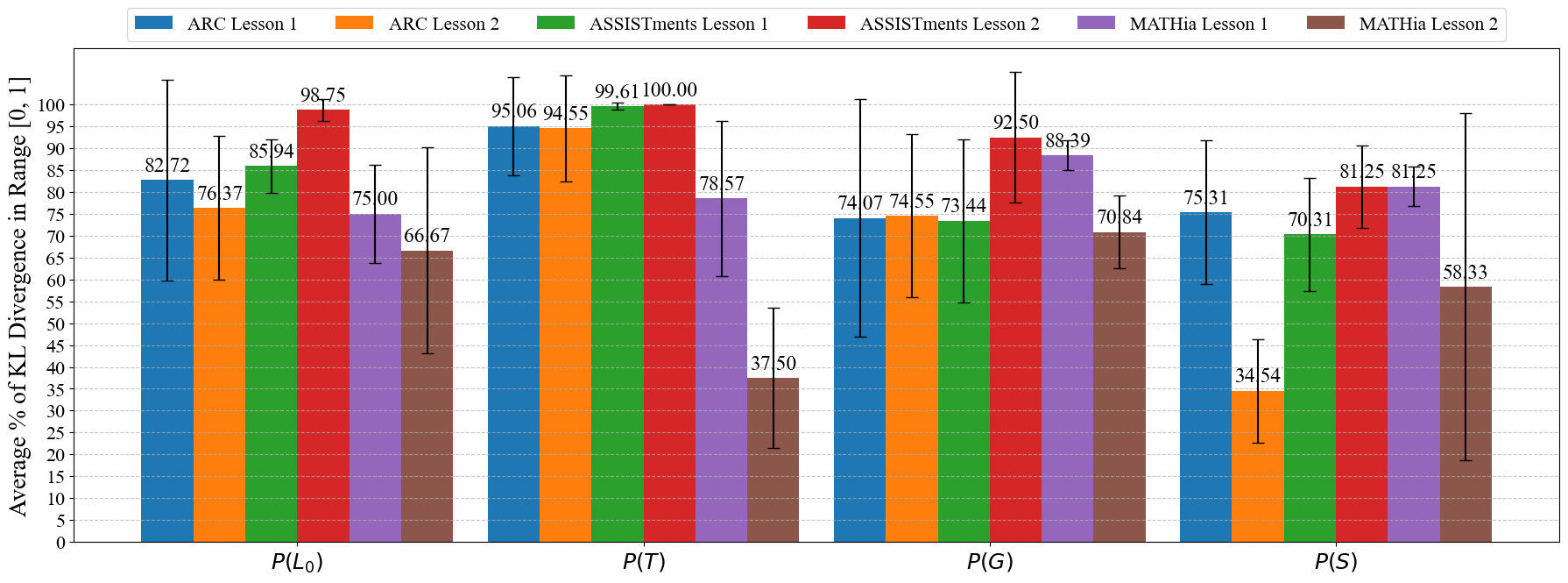}
\caption{Percentages of KL divergence values within [0, 1] for learning parameters derived from BKT modeling.}
\label{fig:kj_comparison_all}
\end{figure*}

The KL divergence measurements across the four learning parameters (\(P(L_0)\), \(P(T)\), \(P(G)\), and \(P(S)\)) encompassing all six lessons provide valuable insights into the alignment between imputed and original data. Figure~\ref{fig:kj_comparison_all} displays the the percentages of KL divergence values within the range [0, 1] for each parameter, aggregated across all questions. Using [0, 1] as an acceptable range for smaller divergence values, as suggested in research from other domains \cite{murphy2012machine,rustad2023digital} (in the absence of specific criteria in learning engineering), provides a practical approach given the unbounded nature of KL divergence. Values closer to 0 indicate a close alignment between the imputed and original data, while values approaching or exceeding 1 suggest greater divergence. The results presented in Figure~\ref{fig:kj_comparison_all}, reveal that most lessons, exhibit high percentages of KL divergence within this range, indicating strong alignment between imputed and original data. This alignment is especially notable for the ASSISTments Lessons, where all parameters consistently show values above 70\% for Lesson 1 and Lesson 2, with both ASSISTments Lesson 1 (\(P(T)\))) and Lesson 2 (\(P(L_0)\) and \(P(T)\)) achieving close to 100\% alignment in several parameters. For the ARC Lessons, there is also generally strong alignment, with most parameters showing percentages above 70\%. However, ARC Lesson 2 shows more variability, especially in \(P(S)\) where the percentage drops to 34.54\%, suggesting that the slip parameter in the imputed data diverges more from the original data in this lesson. This variability in ARC Lesson 2 may reflect challenges in capturing certain learning dynamics, such as the tendency to slip, in the imputed data. The MATHia Lessons, on the other hand, show a mixed pattern of alignment. MATHia Lesson 1 displays moderate to high alignment, with all parameters falling above 70\%. However, MATHia Lesson 2 shows more noticeable divergence across parameters, with \(P(L_0)\) blow 70\%, and \(P(T)\) (37.50\%) and \(P(S)\) (58.33\%) falling below 60\%. This variability, particularly in MATHia Lesson 2, suggests that the imputed data may have limitations in accurately capturing certain learning behaviors in the MATHia dataset. 

In conclusion, the KL divergence analysis results show that the imputation model effectively aligns with the original data across most learning parameters, with minimal divergence observed in \(P(L_0)\), \(P(T)\), \(P(G)\) and \(P(S)\) across the majority of questions. While certain parameters, particularly \(P(S)\) and \(P(T)\), exhibit higher divergence for a subset of questions, these deviations likely reflect the challenges faced by our proposed imputation model in handling sparse data. Overall, these results emphasize the GAIN model's effectiveness in filling missing data while preserving the essential learning characteristics of the original dataset. 

\section{Discussion}

This study proposes a systematic imputation framework that integrates multidimensional learner modeling with GenAI models, specifically GAIN, to address the critical issue of data sparsity in ITSs. Evaluated using three types of ITS lesson datasets (AutoTutor ARC, ASSISTments, and MATHia), our proposed GAIN model's imputation accuracy generally outperforms baseline methods, including Tensor Factorization, CPD, BPTF, GAN, InfoGAN, and AmbientGAN, thus addressed \textbf{RQ1}. Its resilient performance across datasets further highlights its generalizability and adaptability to diverse ITS environments (addressed \textbf{RQ1.2}). Moreover, the GAIN model significantly enhances learner modeling accuracy, as evidenced by improved Bayesian Knowledge Tracing (BKT) performance on imputed data compared to original sparse data (directly addressed \textbf{RQ2.1}). This improvement is closely linked to changes in sparsity rates, where higher sparsity increase rates present greater challenges for accurate imputation. Additionally, KL divergence analysis of key learning parameters (including (initial knowledge \(P(L_0)\), learning rate \(P(T)\), guess rate \(P(G)\)) demonstrates close alignment between imputed and original data, preserving critical learning features (answered \textbf{RQ2.2}). Collectively, these findings affirm the effectiveness of the GAIN model in mitigating data sparsity while maintaining while preserving the original learning data characteristics. 

The 3D tensor-based representation facilitates the estimation of missing performance values for previously unattempted questions and attempts by capturing the complex dynamic interactions and similarities across these dimensions. Within this 3D space, GAIN can impute learner performance by considering the contextual information around specific locations and the similarities among attempts, questions, and even learners, to effectively fill the ``learner performance gaps'' across the entire 3D structure. This approach represents a successful application of the classic Rubin’s imputation rule \cite{little2019statistical} within the learning engineering context, inferring missing performance data based on existing and observed learning behaviors for more effective imputation. The efficacy of this 3D framework is also verified with findings from our previous studies \cite{zhang20243dg,zhang2024generative,zhang2024data}. 

The identification of higher imputation accuracy for sparse learning performance across different types of ITS lessons and varying attempts highlights the compatibility of GAIN with the 3D space and its efficacy in imputing learner performance. In the modeling process, GAIN is adapted to accommodate both the input and output shapes of tensor-based learning performance data. Missing data is imputed through multiple CNN layers in GAIN's generator. The hint matrix, generated based on the sparsity distribution of the original learning performance data, provides conditions for imputing the missing data, while the hint discriminator guides the generator for more accurate imputation. This hint mechanism semantically targets the individual sparsity distribution in the learning data, effectively facilitating data imputation for sparse learning performance data in ITSs. Additionally, the training stage considers the matrix relationships between attempts and questions within the ``learner image'' and accounts for similarities across individual learners within that group. This multidimensional exploration of learning performance enables effective model training on existing data and facilitates the prediction of likely missing performance data based on patterns learned by the neural networks in the GAIN model. 

GAIN’s performance in ARC Lesson 1 underscores the broader challenge of applying generative models to datasets characterized by complex cognitive dependencies, such as reading comprehension. The elevated variance and reduced accuracy observed highlight GAIN’s limitations in handling overlapping and interdependent skillsets, including decoding, vocabulary, and inferential reasoning. These interconnected skills amplify the difficulty of accurately imputing missing values, particularly for learners with uneven cognitive profiles. This case underscores the critical need for models to go beyond conventional imputation techniques, incorporating mechanisms that account for nuanced learner variability and multidimensional cognitive processes. Exploring advanced architectures, such as hybrid or context-sensitive models, may provide better alignment with these complexities. Ultimately, this exception emphasizes the importance of tailoring imputation strategies to the unique demands of specific domains within ITS applications, particularly those as cognitively intricate as reading comprehension.  

To assess the impact of the proposed GAIN-based imputation method on learner modeling, particularly with BKT, we observed a significant improvement in learning modeling accuracy. By reducing data sparsity, the GAIN model provides richer and more complete data, which enhances our understanding of learners’ progress across questions and attempts. This additional information from imputed data allows for a more detailed capture of learners’ performance patterns, leading to more accurate BKT parameter estimates and a fine-grained understanding of key learning features such as initial knowledge, learning rate, guess rate, and slip rate. As a result, the modeling accuracy for imputed data significantly improves, providing a closer alignment with actual learning dynamics compared to models based solely on sparse data. Additionally, this improvement in BKT modeling accuracy is closely tied to changes in sparsity rates as the number of Max Attempt increases. As sparsity rates rise sharply with each additional attempt, the proportion of missing data escalates, making it increasingly complex to accurately impute missing values. This pattern suggests that as students engage in more attempts, maintaining high imputation accuracy becomes more challenging due to the growing scarcity of data points necessary for reliable modeling. Implicitly, sparsity rates seem to be linked to information loss in learning performance, with changes in these rates potentially affecting imputation precision and the overall quality of learner modeling. Further research is needed to clarify this relationship and explore strategies to mitigate the effects of high sparsity rates.

A comparative analysis leveraging KL divergence measurements of the distributions of question-wise learning parameters (initial knowledge \(P(L_0)\), learning rate \(P(T)\), guess rate \(P(G)\), and slip rate \(P(S)\)) between the original sparse and imputed data provides a nuanced understanding of how data imputation affects learning features. This analysis reveals the extent to which the imputed data aligns with the original data, offering insights into the accuracy of the imputation model in preserving the underlying learner behaviors. The generally low KL divergence values, particularly the high percentage within the [0, 1] range, indicate that the imputed data closely mirrors the distributions in the original dataset for all parameters. These results suggest that GAIN-based imputation effectively captures learner dynamics without introducing substantial bias, making it a robust tool for handling missing data in ITS environments. However, some parameters, such as \(P(S)\) and \(P(T)\), display higher divergence in a few specific lessons, potentially reflecting unique or complex learning patterns that the model does not fully capture. Such divergence may arise in lessons where slipping and learning rates vary significantly among learners, or in cases where data sparsity is more pronounced. Future work could explore further refinement of the imputation process, particularly for parameters sensitive to individual variations or specific instructional contexts, as this extends beyond the current study's scope.  

\section{Limitations}

Some misalignment of learning features derived from the imputed dataset may occur, and bias introduced by the imputation process remains a potential concern. Further research is necessary to identify the specific sources of bias resulting from the imputation model and to develop effective methods for mitigating these issues. Additionally, understanding the broader implications of this bias on generalizability and model performance will require more in-depth analysis. The refinement of knowledge components within questions, as well as their connections across different questions, is still an area that needs improvement. This limitation affects the model’s ability to accurately track knowledge transfer and interdependencies between various learning tasks. Moreover, the potential utilization of individual levels of attempts, rather than a one-size-fits-all approach in constructing the 3D tensor of learning performance data, requires further refinement. Although this method offers a theoretically predicted knowledge state in learner modeling for ITSs, the use of equal fixed attempts contributes to increased sparsity levels, necessitating more nuanced methods for managing sparsity and improving precision. Lastly, the varying, and in some cases high, levels of sparsity within the learning performance dataset, as highlighted in Figure~\ref{fig:sparsity}, can significantly impact modeling and analysis, leading to challenges such as model bias, reduced performance, and complexities in knowledge tracing within ITSs. These limitations underscore the need for future research to address the trade-offs between sparsity management, model accuracy, and scalability in educational settings. 

\section{Future Works}

Future research can extend generative data imputation on binary numerical values to dialogue-based scenarios, such as dialogue-based ITSs, leveraging GenAI models like Large Language Models (LLMs). These models could enhance the ability to predict learners' responses, making the imputation process more dynamic and context-aware in dialogue interactive environment. In addition to improving computational models, future work should focus on integrating knowledge entities that are actively engaged in tutoring environments. This would allow simulated learner agents to not only model computational behavior but also reflect actual learning processes more accurately. Sequential question generation and recommendation, tailored to learners' knowledge progression, is another promising area of research. Developing systems that can recommend questions based on a learner’s current knowledge state, and adjusting in real-time, could significantly improve personalized learning. Finally, addressing the impact of information loss on imputation models is crucial. Future work should explore the effects of spatial data distribution on imputation performance, ensuring that both temporal and spatial patterns are effectively captured to reduce information loss and enhance imputation accuracy. 

\section{Conclusions}

In this study, we present a generative data imputation based on GAIN to impute sparse learner performance data in ITSs. By reconstructing the learner performance data into a 3D tensor encompassing learners, questions, and attempts, our method leverages existing data structured in a multidimensional format and adapts along the attempts dimension to accommodate varying levels of sparsity. Enhanced by incorporating convolutional neural networks in the input and output layers and employing a least squares loss function for optimization, our GAIN-based approach aligns the shapes of input and output with the dimensions of the question-attempt matrices along the learners’ dimension. The results of our study shows GAIN gains effective in achieving high-accuracy data imputation, as demonstrated through comparisons with other baselines across three types of ITS lessons including AutoTutor ARC, ASSISTments, and MATHia, while maintaining a close alignment with original data patterns across most learning parameters, as verified by low KL divergence. To assess the impact of data imputation on learning features, we employ BKT modeling to estimate population-level learning parameters. Our analysis of question-wise parameters derived from BKT reveals improved model fitting with imputed data compared to the original sparse data, evidenced by generally low divergence values, suggesting that GAIN preserves essential learning features with minimal bias. However, as sparsity rates increase (especially with additional attempts) the imputation task becomes more complex, and slightly higher divergence in parameters like slip and learning rates in specific lessons indicates areas for refinement to capture unique learning dynamics in high-sparsity conditions. These findings underscore the potential of GAIN-based imputation to enhance learner modeling accuracy, making it a robust tool for supporting adaptive, personalized instruction in ITSs. Future research could explore adaptive modifications within GAIN to improve performance for more complex or individualized learning contexts. Overall, our generative data imputation method effectively overcomes learning performance data sparsity challenges in intelligent  systems for AI education. 

\section*{Acknowledgments}

The research reported here was supported by the Institute of Education Sciences, U.S. Department of Education, through Grant R305A200413 to the University of Memphis. The opinions expressed are those of the authors and do not represent views of the Institute or the U.S. Department of Education. We extend our gratitude to Prof. Mohammed Yeasin from the University of Memphis for his invaluable suggestions on the divergence measurement method employed in this study. Additionally, we sincerely thank Dr. Felix Havugimana from the University of Memphis for his insightful ideas that significantly improved this methodology. 

\bibliographystyle{IEEEtran}
\bibliography{references.bib}
\section*{Biography Section}
\vspace{-40pt}
\begin{IEEEbiography}[{\includegraphics[width=1in,height=1.25in,clip,keepaspectratio]{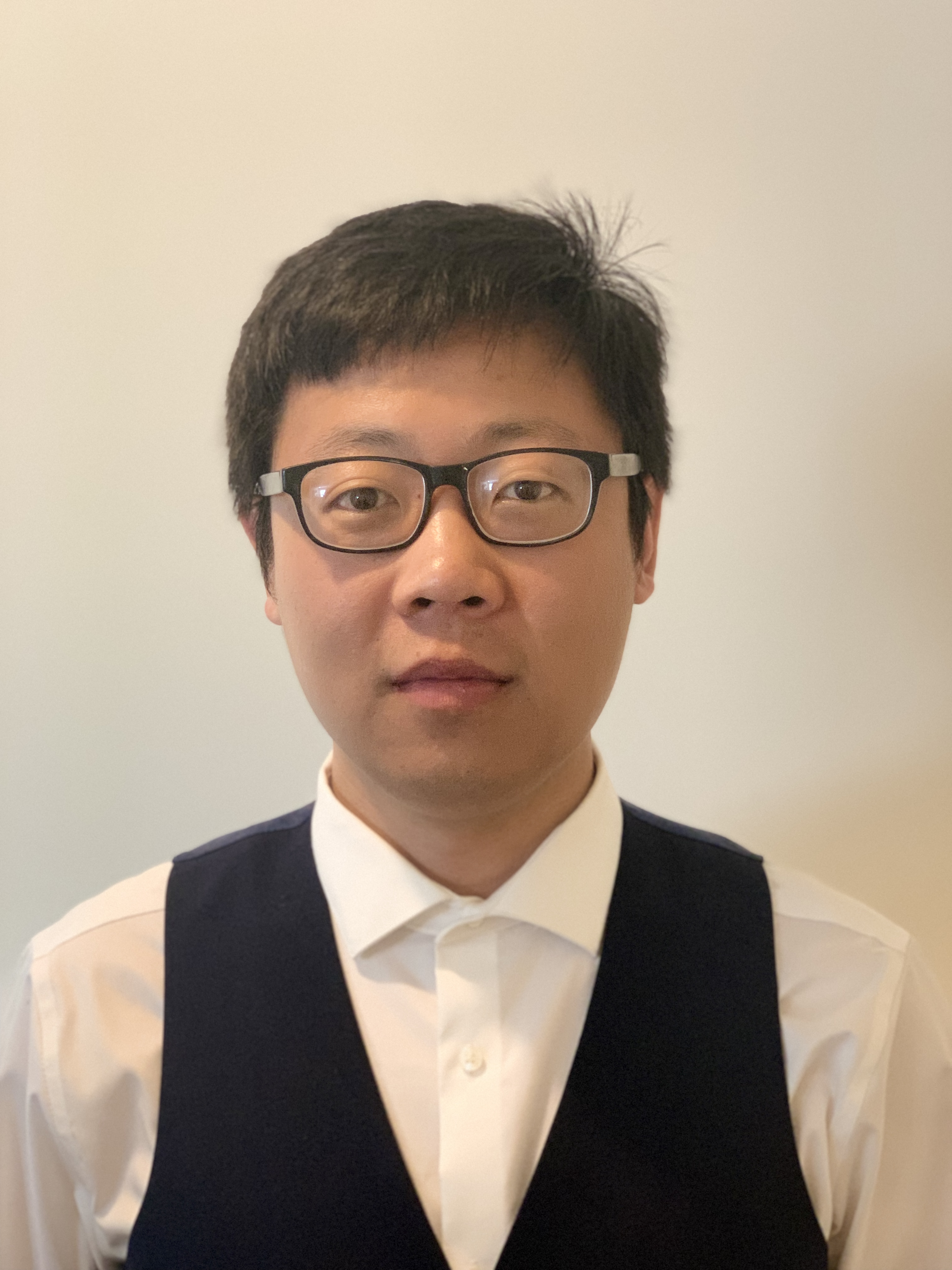}}]{Liang Zhang} is a PhD candidate in Computer Engineering at the Department of Electrical and Computer Engineering, University of Memphis. He is a research assistant at the Institute for Intelligent Systems and currently a visiting scholar at Educational Testing Service in Princeton, NJ. He was an intern researcher at the Human-Computer Interaction Institute at Carnegie Mellon University from May to August 2023, and at the machine learning group at NEC Laboratories America from May to August 2024. His main research interests lie in human-computer interaction, including learning analytics, educational data mining, and Generative AI models, particularly in data imputation and data augmentation within AI-education. He got the Best Full Paper Award at the HCII’24. 
\end{IEEEbiography}
\vspace{-30pt}
\begin{IEEEbiography}[{\includegraphics[width=1in,height=1.25in,clip,keepaspectratio]{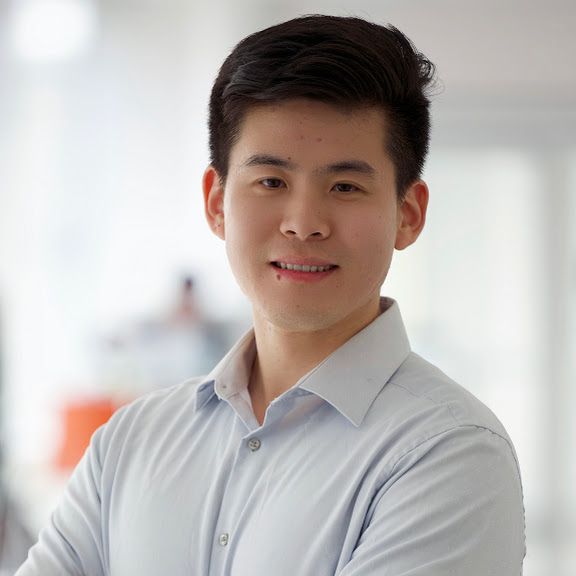}}]{Jionghao Lin} is is an Assistant Professor in the Faculty of Education at the University of Hong Kong. Previously, he served as a Postdoctoral Researcher in the Human-Computer Interaction Institute at Carnegie Mellon University, Pittsburgh, PA, USA (2023–2024). He earned his Ph.D. in Computer Science from Monash University, Clayton, VIC, Australia, in 2023. His research interests include learning science, natural language processing, data mining, and applications of generative artificial intelligence in education. His work has been published in international journals and conferences and recognized with prestigious awards, including the Best Paper Award at HCII'24, EDM'24, and ICMI'19, and the Best Demo Award at AIED'23. 
\end{IEEEbiography}
\vspace{-30pt}
\begin{IEEEbiography}[{\includegraphics[width=1in,height=1.3in,clip,keepaspectratio]{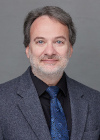}}]{John Sabatini} is a Distinguished Research Professor in the Department of Psychology and the Institute for Intelligent Systems at the University of Memphis. He was formerly a Principal Research Scientist in the Center for Global Research, Research \& Development Division at Educational Testing Service in Princeton, NJ. His research interests and expertise are in reading literacy development and disabilities, assessment, cognitive psychology, and educational technology, with a primary focus on adults and adolescents. He has been the principal investigator of Institute of Education Sciences funded grants to develop pre-K-12 comprehension assessments, as part of the Reading for Understanding initiative, and to adapt those assessments for use in adult education programs, as well as co-PI on a grant project that explores how online collaborative, critical discussions can facilitate the writing of arguments in middle grades students.   
\end{IEEEbiography}
\vspace{-30pt}
\begin{IEEEbiography}[{\includegraphics[width=1in,height=1.3in,clip,keepaspectratio]{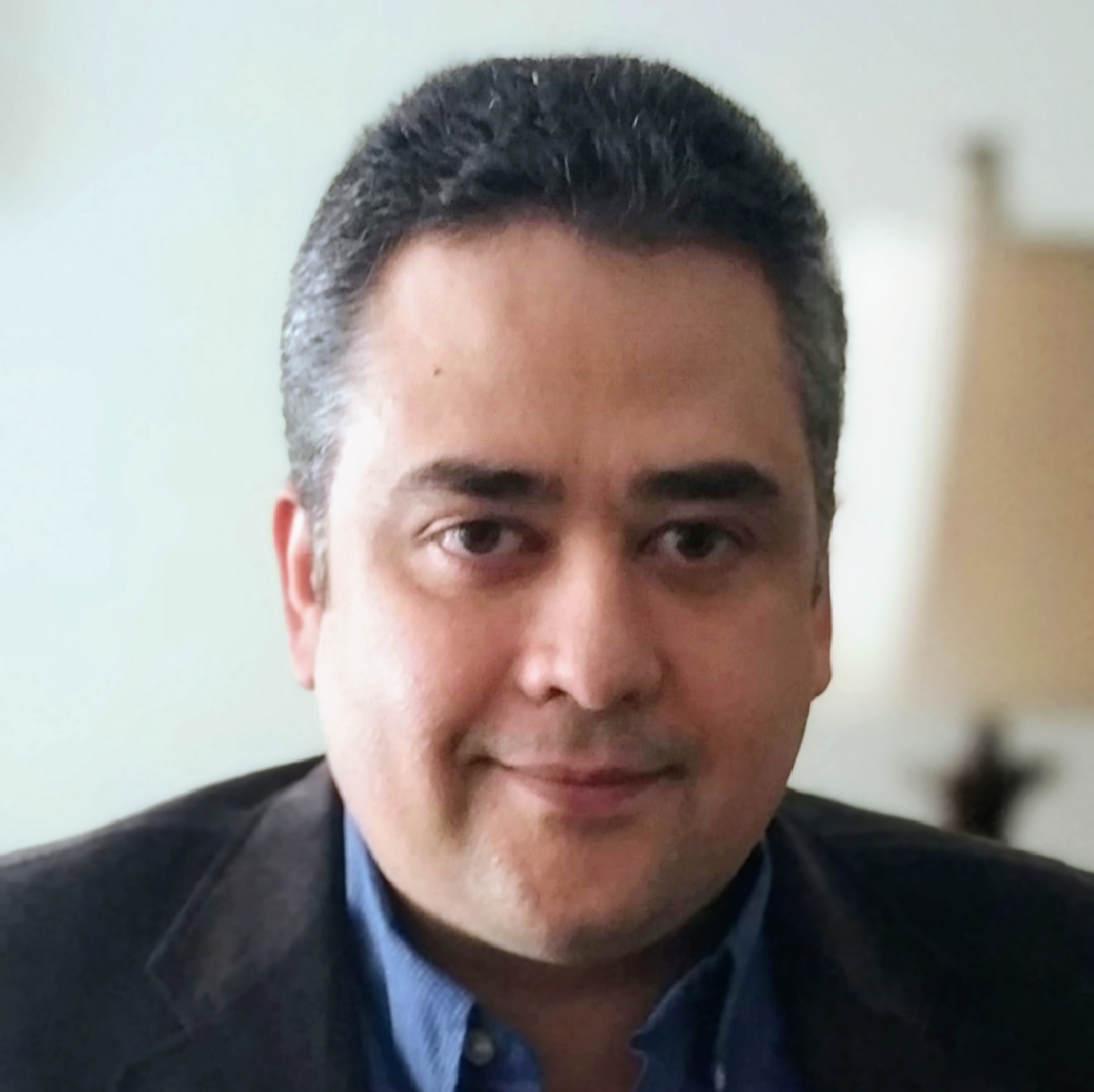}}]{Diego Zapata-Rivera} is a Distinguished Presidential Appointee at Educational Testing Service Research Institute in Princeton, NJ. He earned a Ph.D. in computer science (with a focus on artificial intelligence in education) from the University of Saskatchewan in 2003. His research at Educational Testing Service has focused on the areas of innovations in communicating assessment results to various audiences and technology-enhanced assessment including work on personalized learning and assessment, conversation-based assessment, caring assessment, and game-based assessment.  He has produced more than 150 publications including edited volumes, journal articles, book chapters, and technical papers. Dr. Zapata-Rivera was elected as a member of the International AI in Education Society Executive Committee (2022-2027). He is a Co-PI and research co-director of an NSF AI Institute – the INVITE institute (invite.illinois.edu). He is a member of the Editorial Board of User Modeling and User-Adapted Interaction, Associate Editor for IJAIED, AI for Human Learning and Behavior Change, and former Associate Editor of the IEEE Transactions on Learning Technologies Journal. He is a 2024 IEEE Education Society Distinguished Lecturer.   
\end{IEEEbiography}
\vspace{-30pt}
\begin{IEEEbiography}[{\includegraphics[width=1in,height=1.25in,clip,keepaspectratio]{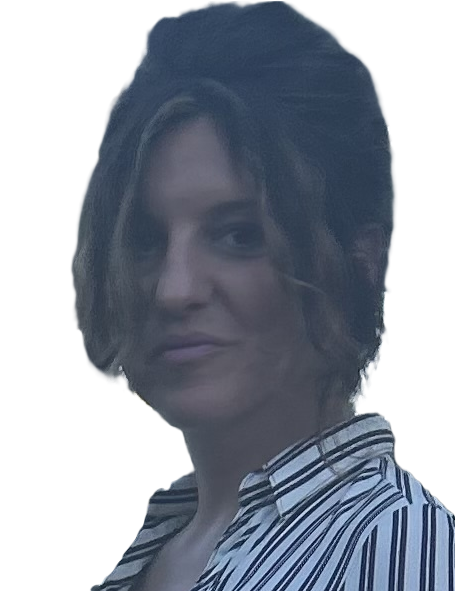}}]{Carolyn Forsyth} is currently a Research Scientist at the Educational Testing Service Research Institute working on novel innovations for digital assessments. She earned her Ph.D. in Experimental Psychology with Cognitive Science Graduate Certification from the University of Memphis in 2014. She has focused her research on understanding and creating environments with natural language conversations for both learning and assessment with a current focus on prompt engineering and development of multi-agent computing systems for generative AI. She also has developed the methodology of theoretically grounded educational data mining to incorporate principled approaches to understanding students processes during interactions with these systems. She has been a key member serving in various roles for both the International Conference on Artificial Intelligence in Education Conference as well as the International Conference on Educational Data Mining for nearly a decade. Furthermore, Dr. Forsyth has contributed over 100 peer-reviewed publications and presentations on this work.  
\end{IEEEbiography}
\vspace{-30pt}
\begin{IEEEbiography}[{\includegraphics[width=1in,height=1.25in,clip,keepaspectratio]{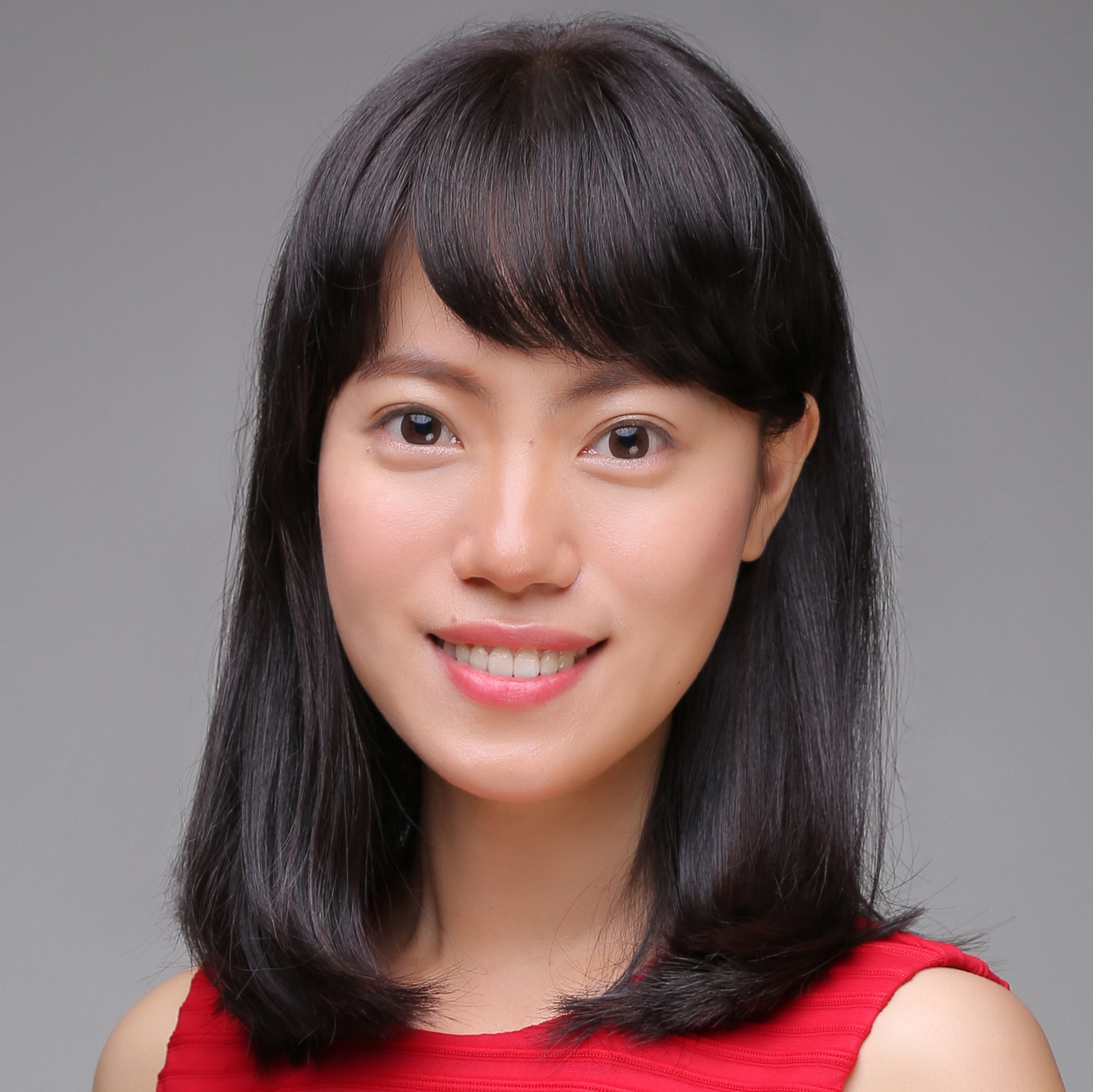}}]{Yang Jiang} a Research Scientist at the Educational Testing Service Research Institute in Princeton, NJ. She received the Ph.D. degree in cognitive science in education from Columbia University, New York, NY, USA, in 2018. Her research focuses on developing and applying AI and data science methods to explore how learners interact with technology-based learning and assessment systems, and how the interactions relate to cognition and learning.    
\end{IEEEbiography}
\vspace{-30pt} 
\begin{IEEEbiography}[{\includegraphics[width=1in,height=1.25in,clip,keepaspectratio]{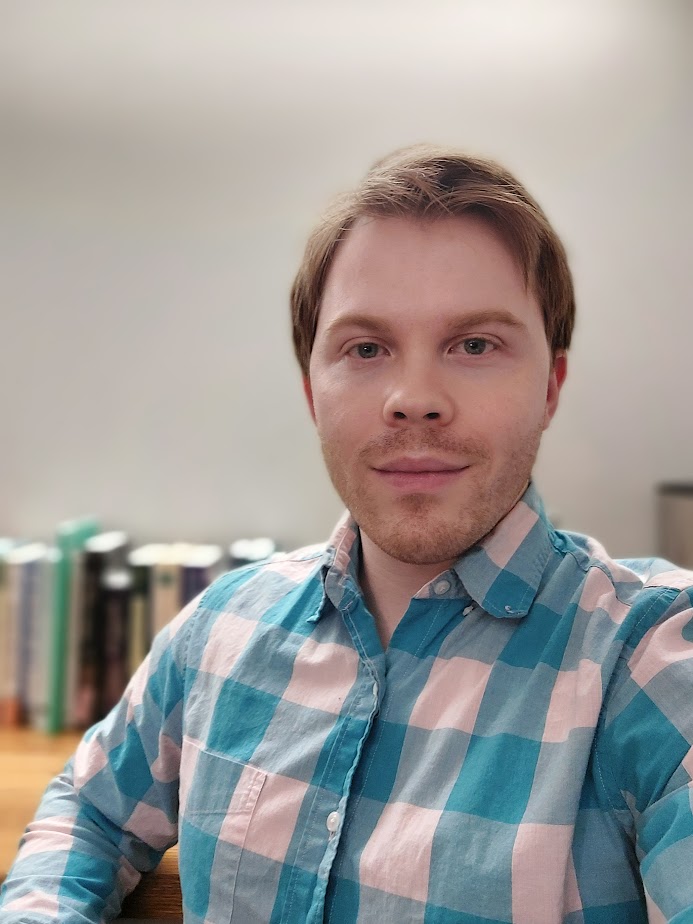}}]{John Hollander} is an Assistant Professor in the Department of Psychology and Counseling at Arkansas State University. His research interests include psycholinguistics, computational semantics, and the science of reading. 
\end{IEEEbiography}
\vspace{-30pt}
\begin{IEEEbiography}[{\includegraphics[width=1in,height=1.25in,clip,keepaspectratio]{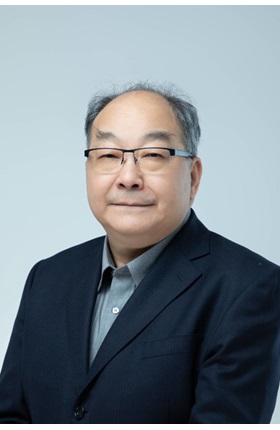}}]{Xiangen Hu} is Chair Professor of Learning Sciences and Technologies at Hong Kong Polytechnic University. He received his Doctor of Philosophy degree in Cognitive Psychology from the University of California, Irvine in 1991 and 1993 respectively. He took up an Assistant Professorship at The University of Memphis in 1993 and was promoted to Associate Professor in 2000 and Professor in the Department of Psychology in 2009. Meanwhile, he has been appointed Prof and Dean of the School of Psychology in the Central China Normal University since 2016 on a visiting basis where he has established good connections with the Chinese mainland. He joined PolyU in Dec 2023 as Director of the Institute for Higher Education Research and Development. His research areas include Mathematical Psychology, Research Design and Statistics, and Cognitive Psychology. 
\end{IEEEbiography}

\vspace{-30pt}
\begin{IEEEbiography}[{\includegraphics[width=1in,height=1.25in,clip,keepaspectratio]{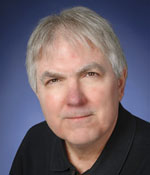}}]{Arthur C. Graesser} is a Distinguished University Professor in the Department of Psychology and the Institute of Intelligent Systems at the University of Memphis and is an Honorary Research Fellow in the Department of Education at the University of Oxford. He received his Ph.D. in psychology from the University of California at San Diego. Art's primary research interests are in cognitive science, discourse processing, and the learning sciences. More specific interests include knowledge representation, question asking and answering, tutoring, text comprehension, inference generation, conversation, reading, problem solving, memory, emotions, computational linguistics, artificial intelligence, human-computer interaction, and learning technologies with animated conversational agents. 
\end{IEEEbiography}



\twocolumn[
\appendix

\section*{Appendix A: Example Prompt Strategy for Augmenting Sparse Learning Performance Leveraging GPT-4}
\label{appendix:A}
]

\begin{figure} [ht!]
\includegraphics[width=0.5\textwidth]{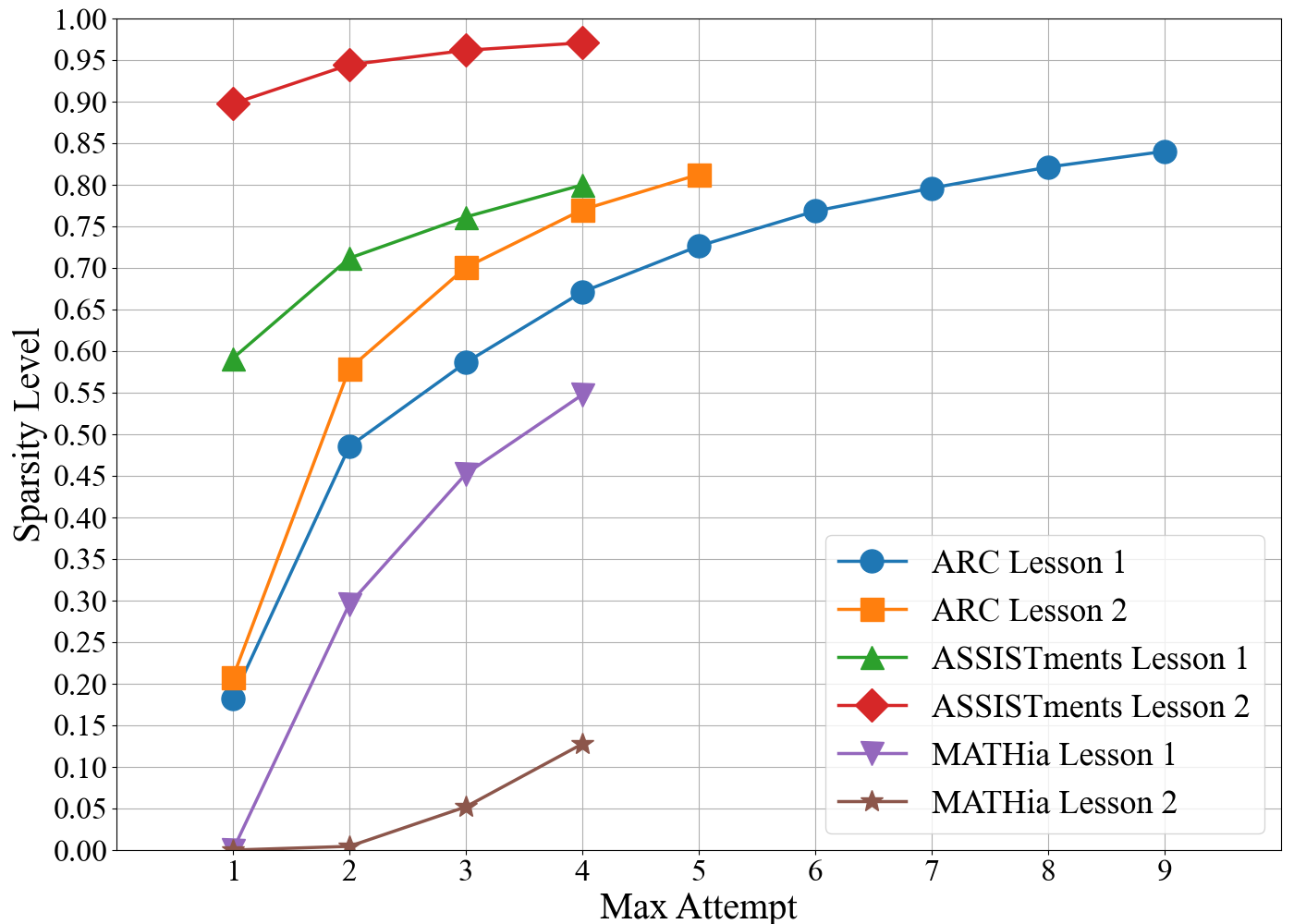}
\centering
\caption{Sparsity Levels Across Different Lessons and Datasets Based on Max Attempt.} \label{fig:sparsity}
\end{figure}

Figure~\ref{fig:sparsity} displays the variation in sparsity levels within learning performance data across six lessons, categorized by the maximum number of attempts. Each line represents a different lesson, with data points showing an increase in sparsity as the number of attempts progresses, suggesting a progressive introduction of missing data or non-responses during the learning process. This trend is consistent across all courses, albeit with varying rates of increase. Notably, ``ASSISTments Lesson 2'' exhibits a gradual ascent, recording the highest sparsity levels across all attempts when compared to other lessons. In contrast, ``MATHia Lesson 1'' and ``MATHia Lesson 2'' demonstrate lower initial sparsity levels, with the former experiencing a sharp increase and the latter following a more gradual trajectory as attempts progress. Particularly, ``ARC Lesson 1'' records the maximum number of attempts observed for this class. The distinct sparsity patterns observed in Figure~\ref{fig:sparsity} highlight the heterogeneity of data completeness and the extent of missingness across different lesson datasets. The distinct sparsity patterns underscore the heterogeneity of data completeness and the extent of missingness across different lesson datasets, which impacts the quality of modeling learners' knowledge states and the ITS's ability to provide accurate, personalized instruction.  

\vspace{10pt}

\clearpage

\vfill

\end{document}